\pdfoutput=1

\documentclass[11pt]{article}

\usepackage[preprint]{acl}

\usepackage{times}
\usepackage{latexsym}
\usepackage{float}

\usepackage[T1]{fontenc}

\usepackage[utf8]{inputenc}

\usepackage{microtype}

\usepackage{inconsolata}

\usepackage{graphicx}

\usepackage[utf8]{inputenc} 
\usepackage[T1]{fontenc}    
\usepackage{hyperref}       
\usepackage{url}            
\usepackage{booktabs}       
\usepackage{amsfonts}       
\usepackage{nicefrac}       
\usepackage{microtype}      
\usepackage{xcolor}         
\usepackage{multirow,multicol}
\usepackage{graphicx}
\usepackage{arydshln}
\usepackage{amsmath}
\usepackage{enumitem}

\usepackage{subcaption}
\usepackage{stmaryrd}

\title{Deriving Character Logic from Storyline as Codified Decision Trees}
 
\author{Letian Peng, Kun Zhou, Longfei Yun, Yupeng Hou, Jingbo Shang \\
University of California, San Diego \\
  \texttt{\{lepeng, kuzhou, loyun, yphou, jshang\}@ucsd.edu}
  }

\usepackage{xspace}

\begin{document}
\maketitle
\begin{abstract}
    Role-playing (RP) agents rely on behavioral profiles to act consistently across diverse narrative contexts, yet existing profiles are largely unstructured, non-executable, and weakly validated, leading to brittle agent behavior. 
We propose \textbf{Codified Decision Trees (CDT)}, a data-driven framework that induces an executable and interpretable decision structure from large-scale narrative data. 
CDT represents behavioral profiles as a tree of conditional rules, where internal nodes correspond to validated scene conditions and leaves encode grounded behavioral statements, enabling deterministic retrieval of context-appropriate rules at execution time. 
The tree is learned by iteratively inducing candidate scene–action rules, validating them against data, and refining them through hierarchical specialization, yielding profiles that support transparent inspection and principled updates. 
Across multiple benchmarks, CDT substantially outperforms human-written profiles and prior profile induction methods on $85$ characters across $16$ artifacts, indicating that codified and validated behavioral representations lead to more reliable agent grounding.\footnote{Codes and datasets used in experiments are available at \href{https://github.com/KomeijiForce/Codified_Decision_Tree}{https://github.com/KomeijiForce/Codified\_Decision\_Tree}}
\end{abstract}

\section{Introduction}

LLM-based Role-playing (RP)~\citep{survey_rp,chenpersona}, (i.e., building established characters into LLMs) focuses on \textit{``How LLMs can interact in a preferred way''}~\citep{llm_human_interaction}, and exhibits broaden applications spanning from emotional support~\citep{SweetieChat}, creative writing~\citep{long_form_story_reasoning}, and gaming engines~\citep{RPGBenchmark}.
Fundamentally, an RP system takes a scene as the input and outputs an action, aiming to imitate the target character's behavior pattern. 
Character profile drives it by grounding the LLM with character-specific information. 
\textbf{Codified profile}~\citep{codification} converts symbolic rules (e.g., \textit{werewolf transforms into wolf under the full moon}) to executable functions (\textit{state = ``wolf'' if is\_full\_moon(scene) else ``human''}). Such a structured design provides explicit, executable constraints, leading to better consistency and interpretability than static textual profiles.

\begin{figure}
    \centering
    \includegraphics[width=0.99\linewidth]{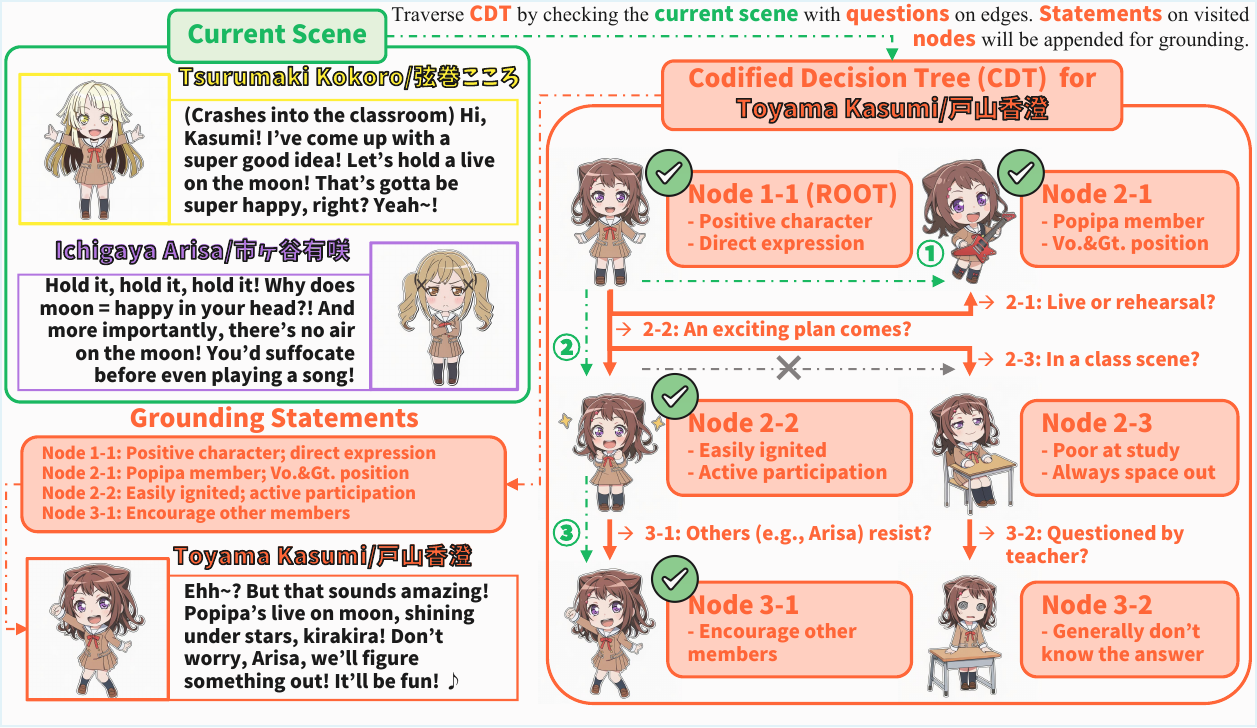}
    \caption{Grounding RP by Codified Decision Tree.}
    \label{fig:kasumi-case}
\end{figure}

Since manually specified profiles are often unavailable or prohibitively expensive to construct, an alternative is to build character profiles from storylines (profiling) automatically. 
Existing profiling methods~\cite{coser} predominantly focus on constructing textual profiles by recurrently or aggregatively merging sub-profiles summarized from blocks of storylines, which often conflates behaviors exhibited under different situational contexts. To address this limitation, we introduce a framework that directly induces codified profiles from storylines, enabling systematic, situation-specific grounding for scene-level action inference. Specifically, inspired by the structure of decision trees (DTs) and human profiling practices, we propose a novel \textbf{Codified Decision Tree (CDT)} method. The core idea is to let the LLM hypothesize candidate scene$\shortrightarrow$action triggers from similar (scene, action) pairs in the storyline, and then validate these triggers against the complete set of observed pairs.

Concretely, we construct CDT by inducing a recursively defined tree of behavioral rules from observed (scene, action) pairs. We first cluster semantically similar pairs using text embeddings to surface candidate scene$\shortrightarrow$action regularities. Within each cluster, an LLM proposes codified triggers of the form \textit{``if A then B''} that map interpretable situational predicates $A$ to actions or action modifiers $B$. These hypothesized triggers are then systematically validated against the entire dataset and used to grow the tree recursively: highly predictive triggers are promoted to internal nodes that partition the data, unsubstantiated triggers are discarded, and partially predictive triggers are refined into child subtrees over filtered subsets of pairs. This construction process yields a hierarchy in which each node stores a compact set of codified grounding statements, and edges correspond to discriminative questions about whether a scene satisfies the associated conditions.

At inference time, a novel scene is routed through the CDT by answering these questions, going through edges whose question return \textit{``True''}. Statements on nodes along the visited path form its situation-specific grounding information, which is then provided to the RP policy for action generation. The resulting structure supports transparent inspection and editing of behavioral rules, principled incorporation of additional data via local subtree updates, and deterministic retrieval of context-appropriate codified profiles for diverse scenes. An example for CDT traversal is presented in Figure~\ref{fig:kasumi-case}, where our protagonist \textit{``Toyama Kasumi''} (from \textit{``BanG Dream! Project''}) is facing a scene inviting her to hold a live on the moon. Kasumi’s CDT works by following only the edges with questions satisfied by the scene, accumulating statements (easily ignited and vocalist position) from visited nodes and deliberately omitting ones from unvisited alternatives (study-related branches), which demonstrates the situation-aware nature of CDT.

To test the performance of CDT, we first adapt the existing RP benchmark~\citep{codification} to a finer-grained one through all action extraction and action sequence modeling.
Then, we enrich it by adding new artifacts, curating new styles, and collecting large-scale event story conversations.
In total, we have $45$ characters, $20,778$ scene-action pairs in the fine-grained Fandom benchmark, $40$ characters, $7,866$ pairs in the Bandori benchmark, and an extra $77,182$ pairs of event story conversations from \textit{``BanG Dream!''}.
Based on these benchmarks, we compare CDT against alternative ways of leveraging the training data, including model fine-tuning, retrieval-based in-context learning, and prior textual profiling methods, and find that CDT yields substantially stronger action-prediction performance. Notably, CDT even surpasses human-written profiles, highlighting the effectiveness of data-driven profiling when coupled with codification for situation-specific grounding.

We further conduct an in-depth analysis of CDT by examining both its profiling and traversal configurations, including hyperparameter choices, efficiency, clustering mechanisms, and top-$k$ statement selection strategies. In addition, we build a conversion pipeline that translates CDT into a reader-friendly wiki-style textual profile, 
and the resulting profiles still outperform human-written ones, underscoring the intrinsic quality of CDT-induced knowledge. We also study the data-driven scaling-up of CDT, showing that more data results in stronger CDT profiling. Finally, we present relation modeling as a case study of goal-driven CDT, where the tree is constructed to emphasize a specific behavioral aspect of a character, demonstrating that such targeted profiling can further improve RP precision. Our contributions are three-fold:

\begin{itemize}[nosep,leftmargin=*]
    \item We propose Codified Decision Trees (CDT), a novel data-driven algorithm that induces executable, situation-specific character profiles from storylines for grounded RP inference;
    \item We introduce enriched benchmarks and conduct comprehensive evaluations, concluding that CDT consistently outperforms other profiling approaches, and even human-written profiles;
    \item We provide an in-depth analysis of CDT, including configuration setup, training and inference strategies, scaling up, and converting CDT to in-depth textual profiles.
\end{itemize}
\section{Background}

\paragraph{Role-playing.}
Role-playing (RP) tasks aim to sustain coherent and persona-consistent behaviors across evolving narrative or simulated environments~\citep{riedl2012interactive,shao2023character,chenpersona}. 
Early RP systems relied on hand-crafted character sheets and rules that describe goals, traits, and responses, but these were often limited in scope and lacked systematic validation.
Recent advances in large language models (LLMs) have enabled richer, adaptive personas~\citep{yan2023larp,moore2024rolellm,apc}, where characters can recall, reason, and act through long-horizon interactions.
However, maintaining behavioral stability and transparency remains challenging, because profiles represented as plain text are hard to verify or execute, and also prone to cause inconsistency. 
Structured representations such as reasoning graphs, hierarchical memories, and codified constraints~\citep{cheng2025psymem,codification,tang2025thinking} have been proposed to stabilize persona behavior, but these often require extensive manual definition.
Our work builds on this direction by inducing interpretable, executable behavioral structures directly from narrative data, enabling grounded, deterministic role-play without handcrafted profiles.

\paragraph{Grounding System.}
Grounding connects agents’ internal reasoning to external narrative or environment states, ensuring that actions and dialogue remain contextually valid~\citep{liang2022code,wu2023tidybot}. 
In RP and simulation contexts, grounding allows to condition agent responses on evolving world facts, social relations, and temporal cues rather than free-form text history.
Prior methods achieve this via world models~\citep{zhang2024grounding,liu2024grounded}, graph-based memory~\citep{li2024graphreader}, or structured symbolic stores that track entities and events~\citep{sun2024identity}.
These methods improve consistency but often rely on manually crafted schemas or domain-specific rules.
In contrast, we automatically construct codified decision trees from large-scale narrative corpora. The induced structure explicitly encodes grounding conditions and decision logic, allowing systematic traversal and verification during execution.

\paragraph{Rule Mining.}
Rule mining seeks to discover conditional regularities, typically expressed as if–then statements that govern observable behaviors in data~\citep{wang2024fsm}. 
Traditional approaches identify statistical or symbolic dependencies between events, while recent LLM-based methods infer and validate such rules through natural language understanding and generation~\citep{gan2024can,yoneda2024statler,wang2025sg}.
These rules have been used for planning~\citep{yao2023react,gao2023pal}, reasoning~\citep{li2022competition}, and social simulation~\citep{sun2024identity}, providing interpretable scaffolds for decision control.
However, most existing work treats rule extraction as a flat, unstructured process and offer limited mechanisms for recursive refinement or contextual specialization.
Our CDT differs by inducing hierarchical, executable rule structures: candidate triggers are mined, validated, and recursively expanded, forming a transparent tree that encodes both coverage and precision for role-specific behaviors.
\section{Codified Decision Tree}

\begin{figure*}
    \centering
    \includegraphics[width=0.9\linewidth]{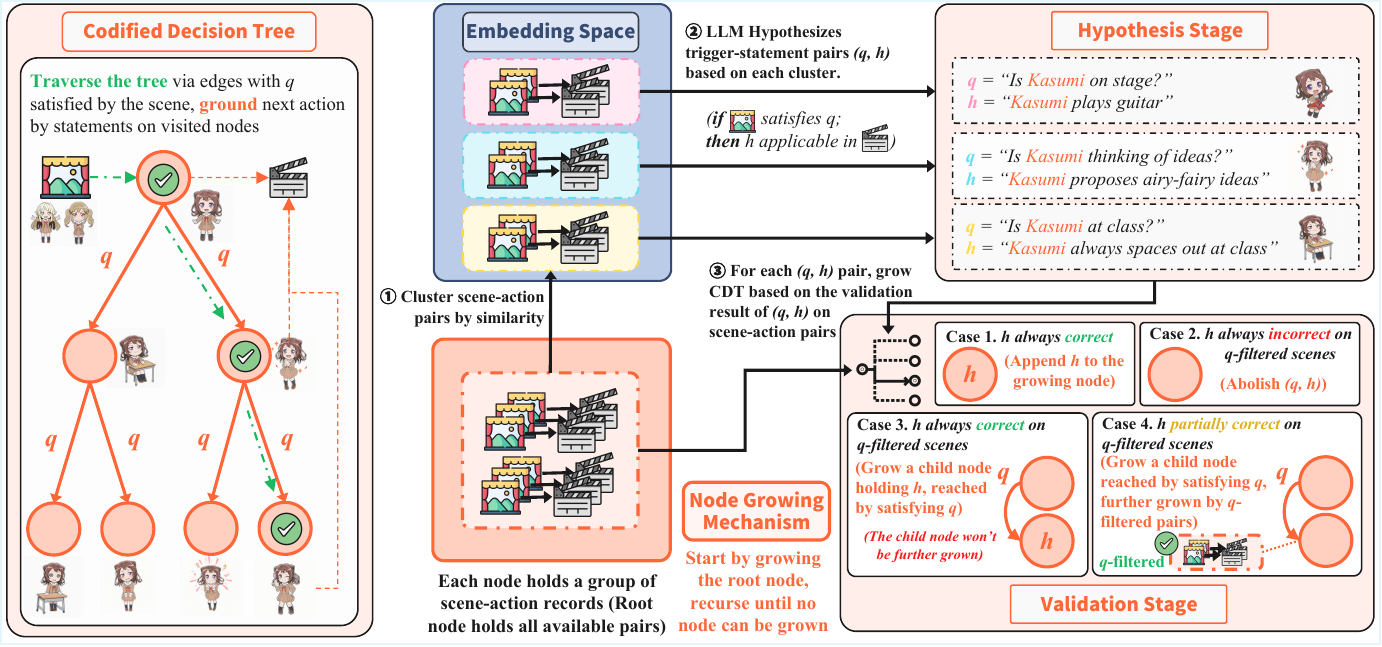}
    \vspace{-2mm}
    \caption{The workflow of codified decision tree (CDT).}
    \label{fig:cdt_main}
    \vspace{-5mm}
\end{figure*}

\subsection{Preliminary}

\paragraph{Role-playing system} It can be formatted as a function $a = \textrm{RP}(s|g_x)$ that takes scenes $s$ as inputs and outputs actions $a$ conditioning on a character $x$'s information $g_x$. As most modern RP systems are driven by LLMs, we replace $\textrm{RP}(\cdot)$ by $\textrm{LLM}_\textrm{RP}(\cdot)$ in this paper. The process to incorporate $g_x$ into the action generation is called grounding. A grounding system can be static (e.g., appending $x$'s profile to the context) or dynamic (e.g., retrieving a similar scene-action pair).

\paragraph{Profile Codification} Among dynamic systems, codification~\citep{codification} is a recently proposed RP grounding paradigm, which converts the static profiles into executable functions $f: s\rightarrow g_x$. For example, a statement in $x$'s profile:

\begin{center}
\small
    \textit{``$x$ is always brave against all kinds of challenges.''}
\end{center}

\noindent will be codified as

\begin{center}
    \small
    \texttt{if check($s$, }\textit{``challenge exists?''}\texttt{) return} \textit{``brave''}
\end{center}

\noindent where \texttt{check($s$, $q$)} is a discriminator calling that returns \texttt{True/False/None} (\texttt{None} represents ``Unknown'') based on the scene $s$ and question $q$. The discrimination can be executed by the RP LLM itself or a fine-tuned model. Such a codification process improves the RP LLM's consistency with the profile logic by enforcing reasoning paths rather than leaving it for LLMs to explain. A limitation of the existing codification framework is the reliance on human-written profiles for conversion. This paper explores a different data-driven codification, which takes a set of scene-action pairs $\mathcal{D}=\{(s_i, a_i)\}_{i=1:|\mathcal{D}|}$ as input to directly build the codified function $f$ without given profiles.

\subsection{Data Structure}

Similar to traditional decision trees, CDTs take a textual scene $s$ as the input and enable moving between nodes $v$ based on discrimination, which is implemented as the mentioned \texttt{check($s$, $q$)} function. The difference is that CDTs do not output a single label like traditional decision trees. As a grounding system, CDTs will instead be traversed, and all grounding information inside visited nodes will be incorporated for grounding. As shown in Figure~\ref{fig:kasumi-case}, a node in CDT includes two key elements: 1) a set $H$ of statements $h$; 2) a set of child nodes with questions for traversal checking.

\paragraph{CDT Traversal} (Shown in Figure~\ref{fig:kasumi-case}, with another running case available in Figure~\ref{fig:running_inference}) starts from the root node with an input scene $s$, and we first append all statements inside the root node to the grounding set $g$, then for each child node, we check whether \texttt{check($s$, $q$)} is \texttt{True} for the assigned question. If the check is passed, we visit the corresponding child node and recursively execute the flow: 1) appending statements; 2) checking to visit child nodes operations until all reachable nodes are visited. 
Follow the proposed structure, each node in CDT can be understood as the triggered behavior statements under conditions filtered by multiple questions $q$. 

\subsection{Recursive Hypothesis-Validation}
\label{subsec:rhvm}

Based on CDT's definition above, we design the methodology to grow a CDT (Figure~\ref{fig:cdt_main}). For initialization, we have a root node with no statements or children, but the whole training set of scene-action pairs $\mathcal{D}_{train}$. Then we apply a rule mining system (elaborated in $\S$~\ref{subsec:rule_mine}) to hypothesize the causality between scenes and actions (hypothesizing case available in Figure~\ref{fig:running_hypothesis}). Each hypothesis is in the format $(q, h)$ where $q$ is the filtering question and $h$ is the behavioral statement.

For each $(s, a)$ inside $\mathcal{D}_{train}$, we first check whether $h$ is a global ($q$-free) behavior statement according to the $h$-$a$ natural language inference (NLI) relation. By NLI discrimination on all data, we get the ratio of entailed $r_e$, neutral $r_n$, and contradicted $r_c$. If the accuracy $\frac{r_e}{r_e+r_c}$ reaches an accept threshold $\theta_\textrm{acc}$, $h$ will be directly added to the statement set $H$ as a globally applicable behavior.

For $(q, h)$ failed to be established as a global statement, we then filter $\mathcal{D}$ into a subset $\mathcal{D}'$ of $(s, a)$ pairs that $s$ passes the \texttt{check($s$, $q$)}. Based on the accuracy, an accept threshold $\theta_\textrm{acc}$, and a reject threshold $\theta_\textrm{rej}$, we select the next operation step.

\begin{itemize}[nosep,leftmargin=*]
    \item If accuracy $> \theta_\textrm{acc}$, we add a leaf node with only $h$ in $H$, which can be accessed from the parent by \texttt{check($s$, $q$)} and won't be further grown, representing a critical hit.
    \item If accuracy $< \theta_\textrm{rej}$, $(q, h)$ is abolished with no further operation, representing a failed hypothesis.
    \item If accuracy falls between $\theta_\textrm{rej}$ and $\theta_\textrm{acc}$, we view it as a logic that requires further exploration for more complex structures. Thus, we will build a new empty node connected to its parent by \texttt{check($s$, $q$)}, the new node will take the filtered dataset $\mathcal{D}'$ for recursive growth. To avoid recursion lasting too long, we set a filtering threshold $\theta_f$ so that the recursion is only triggered when $\frac{|\mathcal{D}'|}{|\mathcal{D}|} < \theta_f$, which gradually reduces $|\mathcal{D}|$ during recursion. The recursion can also be stopped by a too small $|\mathcal{D}|$ or too deep $v$ in the CDT.
\end{itemize}

By such a recursive hypothesis-validation mechanism, we can model the logic in different complexities to ensure broad and deep coverage. 

\subsection{Rule Mining}
\label{subsec:rule_mine}

In the recursive hypothesis-validation mechanism ($\S$~\ref{subsec:rhvm}), the rule mining system is a crucial component to make high-potential hypotheses for $(q, h)$. 
As we target a more precise and efficient mining of $(s, a)$ pairs that represent the causality, we apply a clustering algorithm (K-Means) based on the textual embeddings of $s$ and $a$. 
For $a$, we apply a semantic textual embedding~\citep{simcse}. For $s$, as we care more about the potential triggered character behavior, we follow the idea of instruction-following embedding~\citep{Inbedder} to use the last token prediction hidden state from a generative LM as the embedding following the prompt below:

\begin{center}
    ``\{scene\} Thus, \{character\} decides to''
\end{center}

\noindent where the hidden states predicted from \textit{``to''} will contain the distribution of all kinds of verbs triggered by $s$ to guide the clustering. In contrast, a normal semantic embedding will be distracted by surface similarity, especially by clustering scenes from the same episode together, ignoring their difference in impact on the target character. Based on clustering the concatenated embedding between scenes and actions, we prompt LLM to summarize potential rules from each cluster. 
\paragraph{Diversification} We let parent nodes pass their question paths (the sequence of questions to reach the node) and established statements (the validated statements when reaching the node) to children. Such information is incorporated into the hypothesis prompt to instruct the LLM to propose something else to avoid redundant checking.
\section{Benchmark}

While early RP works rely on synthesized datasets~\citep{eu,in_character}, the RP community has also begun to use humans' annotations (e.g., profiles and synopses)~\citep{bookworld,coser} on well-known artifacts for benchmarking. Fandom Benchmark~\citep{codification} is such an RP benchmark, which utilizes Fandom\footnote{\href{https://www.fandom.com/}{https://www.fandom.com/}}'s rich human annotations for both profiles and synopses. Fandom Benchmark extracts \textbf{only key actions} of characters in scenes to evaluate whether RP systems can take a given scene to make an action entailed by the ground-truth original actions in NLI. We extend such a benchmarking strategy to \textbf{all actions} in the gathered artifacts.

\begin{table}
\centering
\small
\scalebox{.72}{
\begin{tabular}{lp{8.5cm}}
\toprule
& Haruhi is unworried, pointing out the North High uniform is not distinctive enough to identify at a glance\\
\cmidrule(l){2-2}
\multirow{3}*{Scene} & Haruhi reveals her concept for the film is for the timid Asahina to go through great struggles and suffering to make the happy ending more satisfying\\
\cmidrule(l){2-2}
& Koizumi asks if there will be any actors in the film other than Asahina, Nagato, and himself\\
\midrule
Question & What'll be Haruhi's next action in response to the current scene?\\
\midrule
Action & Haruhi is inspired to recruit people to play Nagato's minions \\
\bottomrule
\end{tabular}
}
\vspace{-2mm}
\caption{Examples of training/test cases. ($10$ preceding actions as the scene in real benchmarks)}
\vspace{-3mm}
\label{tab:efficiency}
\end{table}

\paragraph{Fine-grained Fandom Benchmark}
We use the same source of story synopses crawled from Fandom as the original benchmark. Instead of extracting only key actions in the original benchmark, we utilize an LLM to break narrations into sequences of actions annotated with acting characters (\textit{``environment''} when no active character). The preceding $10$ actions of each action are taken as the input scene. For each character, we perform a chronological split of scene–action pairs, training on the first half of the storyline and evaluating on the second half, which both prevents data contamination and mirrors real-world prediction of future behavior from past evidence. In addition to the original $6$ artifacts {(\textit{``Haruhi''}, \textit{``K-On!''}, \textit{``Fullmetal Alchemist''}, \textit{``JOJO''}, \textit{``A Game of Thrones''}, \textit{``Avatar: The Last Airbender''})}, we incorporate \textit{``Death Note''} and \textit{``Spy $\times$ Family''} to broaden the benchmarking scope. We experiment on the main characters that appear throughout the storyline for long-horizon testing.

\paragraph{Bandori Conversational Benchmark}
We curate the Bandori conversational benchmark for assessing dialogue-level role-playing behavior. We collect conversations from the first band story of all eight bands (from \textit{``Poppin'Party''} to \textit{``MyGO!!!!!''})\footnote{e.g., \href{https://bandori.fandom.com/wiki/Poppin\%27Party/Band_Story}{bandori.fandom.com/wiki//Poppin'Party/Band\_Story/}}, where each utterance is treated as an action. Artifact and character background information is provided in Appendix~\ref{apdx:character_info}.

\paragraph{Criterion and Statistics}
For both benchmarks, predicted next actions are compared against reference actions using reference-prediction NLI relation (\textit{``entailed''} for $100$, \textit{``neutral''} for $50$, \textit{``contradicted''} for $0$), following prior practice~\citep{codification}. Overall, the fine-grained Fandom benchmark contains $45$ characters and $20{,}778$ scene-action pairs, while the Bandori benchmark comprises $40$ characters and $7{,}866$ pairs. We further collect \textit{``BanG Dream!''} event-story conversations, yielding extra $77{,}182$ pairs for scaling-up analysis. Statistics are provided in Appendix~\ref{apdx:stats}. Our main content focuses on the NLI score for next action prediction, while multi-dimensional scoring, out-of-domain scenario, and human evaluation can be found in Appendix~\ref{apdx:extended_metric}, which validates NLI to be cross-metric and human consistent.
\section{Experiment}

\begin{table*}
\centering
\small
\scalebox{.9}{
\begin{tabular}{llccccccccc}
\toprule
\multicolumn{2}{l}{Fandom} & {Haruhi} & {K-On!} & {S$\times$F} & {DN} & {FMA} & {JOJO} & {AGOT} & {ATLA} & {Avg.} \\
\midrule
\multirow{6}*{Data-driven} & Vanilla & $55.08$ & $49.92$ & $56.10$ & $62.49$ & $55.66$ & $54.66$ & $57.05$ & $53.56$ & $55.57$ \\
& Fine-tuning & $51.49$ & $51.01$ & $49.14$ & $50.79$ & $44.07$ & $34.92$ & $41.20$ & $42.84$ & $45.68$ \\
& RICL & $56.83$ & $55.74$ & $56.86$ & $62.80$ & $56.33$ & $49.77$ & $56.46$ & $52.25$ & $56.01$ \\
& ETA & $60.54$ & $53.83$ & $58.00$ & $63.29$ & $57.12$ & $51.00$ & $55.28$ & $56.23$ & $56.91$ \\
& CDT (Ours) & $61.16$ & $57.93$ & $60.35$ & $66.34$ & $58.57$ & $57.40$ & $63.79$ & $61.05$ & $60.82$ \\
& CDT-Lite (Ours) & $62.17$ & $57.24$ & $59.79$ & $67.00$ & $59.04$ & $57.26$ & $64.27$ & $61.32$ & $\textbf{61.01}$ \\
\midrule
\multirow{2}*{Human} & Human Profile & $55.87$ & $55.86$ & $59.14$ & $64.75$ & $58.54$ & $55.11$ & $59.35$ & $57.98$ & $58.33$ \\
& Codified Human Profile & $57.94$ & $55.93$ & $59.38$ & $65.56$ & $57.01$ & $56.56$ & $62.07$ & $59.97$ & $59.30$ \\
\midrule
\midrule
\multicolumn{2}{l}{Bandori} & {PoPiPa} & {AG} & {PasuPare} & {Roselia} & {HHW} & {Monica} & {RAS} & {MyGO} & {Avg.} \\
\midrule
\multirow{6}*{Data-driven} & Vanilla & $66.39$ & $66.76$ & $68.29$ & $66.83$ & $65.13$ & $64.06$ & $67.20$ & $59.37$ & $65.50$ \\
& Fine-tuning & $69.52$ & $62.76$ & $64.63$ & $61.83$ & $62.39$ & $62.35$ & $62.64$ & $56.72$ & $62.86$ \\
& RICL & $73.56$ & $67.56$ & $73.06$ & $67.24$ & $73.63$ & $65.04$ & $69.44$ & $61.10$ & $68.86$ \\
& ETA & $75.29$ & $72.49$ & $78.00$ & $70.91$ & $78.92$ & $66.82$ & $72.68$ & $62.89$ & $72.25$ \\
& CDT (Ours) & $84.25$ & $79.92$ & $78.93$ & $71.93$ & $80.03$ & $77.33$ & $80.08$ & $69.17$ & $77.71$ \\
& CDT-Lite (Ours) & $88.38$ & $80.49$ & $82.47$ & $72.81$ & $79.66$ & $78.67$ & $79.51$ & $70.33$ & $\textbf{79.04}$ \\
\midrule
\multirow{2}*{Human} & Human Profile & $73.73$ & $72.43$ & $77.11$ & $70.08$ & $73.14$ & $68.08$ & $71.74$ & $63.91$ & $71.28$ \\
& Codified Human Profile & $73.02$ & $74.00$ & $78.65$ & $71.23$ & $72.47$ & $69.14$ & $71.41$ & $65.02$ & $71.87$ \\
\bottomrule
\end{tabular}
}
\vspace{-2mm}
\caption{RP performance comparison (NLI score) on fine-grained Fandom and Bandori benchmarks.}
\vspace{-3mm}
\label{tab:main}
\end{table*}

\subsection{Evaluation and Baselines}

We include comprehensive baseline methods to cover all types of existing methods that utilize known plots to ground RP in new scenes.


\begin{itemize}[nosep,leftmargin=*]
    \item \textbf{Vanilla} directly prompts the RP model with the input scene, without any additional information.
    
    \item \textbf{Fine-tuning} adapts the RP model by supervised training on the character-specific scene–action pairs, enabling implicit memorization of behavioral patterns from past storylines.
    
    \item \textbf{Retrieval-based In-Context Learning (RICL)}~\citep{ricl} retrieves a set of scene–action examples from the training data that are most similar to the input scene as in-context examples to guide action generation.
    
    \item \textbf{Extract-Then-Aggregate (ETA)}~\citep{coser} first extracts textual sub-profiles from blocks of storylines and then aggregates them into a single character profile, which is appended to the prompt as grounding information.

    \item \textbf{Human Profile} grounds RP using human-written character descriptions, containing natural language canonical traits and behaviors. Our experiment applies the ones written in Fandom wiki.
    
    \item \textbf{Codified Human Profile}~\citep{codification} converts human-specified symbolic rules into executable grounding functions that condition actions on scene-specific predicates.
\end{itemize}

\subsection{Implementation Details}

\paragraph{Hyperparameters}
Detailed hyperparameter setups are placed in Appendix~\ref{apdx:hyperparameter}. Hyperparameter impact analysis can be found in Appendix~\ref{apdx:extended_analysis}.

\paragraph{Clustering}
We embed actions by using \texttt{qwen3-embedding-8b}~\citep{qwen3_embed}. Scenes are embedded using instruction-following representations produced by the generative model \texttt{qwen3-8b}~\citep{qwen3}.

\paragraph{Hypothesis-Validation}
We use \texttt{gpt-4.1} to hypothesize candidate scene$\rightarrow$action triggers in natural-language ``if–then'' form. Trigger validation is conducted with \texttt{gpt-4.1-mini} via NLI-style judgments over all scene–action pairs. We also introduce a lightweight variant CDT-Lite, by replacing \texttt{gpt-4.1-mini} with a 0.1B encoder model (\texttt{deberta-v3-base}~\citep{debertav3}) distilled from $1\%$ of \texttt{gpt-4.1-mini} discrimination.

\paragraph{RP Model}
\texttt{llama-3.1-8b-instruct} is used as the RP model to generate character actions and responses in main experiments. It is also used to answer discriminative questions during CDT traversal. When using CDT-Lite, question discrimination is instead performed by \texttt{deberta-v3-base}. 

\subsection{Main Results}

\begin{table*}
\centering
\small
\scalebox{.86}{
\begin{tabular}{llccccccccc}
\toprule
\multicolumn{2}{l}{\textbf{Fandom}} & {Haruhi} & {K-On!} & {S$\times$F} & {DN} & {FMA} & {JOJO} & {AGOT} & {ATLA} & {Avg.} \\
\midrule
\multicolumn{2}{l}{CDT-Lite ($d_\textrm{max}=4$)} & $62.17$ & $57.24$ & $59.79$ & $67.00$ & $59.04$ & $57.26$ & $64.27$ & $61.32$ & $61.01$ \\
\midrule
\multirow{7}*{\rotatebox{90}{Ablation}}& $\quad$ w/ Abolished Statements & $61.18$ & $55.09$ & $58.85$ & $66.21$ & $58.01$ & $56.24$ & $61.29$ & $59.33$ & $59.52$ \\
& $d_\textrm{max}=1$ & $60.99$ & $55.75$ & $60.88$ & $66.29$ & $58.03$ & $54.56$ & $61.03$ & $60.43$ & $59.65$ \\
& $d_\textrm{max}=2$ & $61.03$ & $56.61$ & $60.83$ & $68.59$ & $58.82$ & $55.61$ & $62.13$ & $60.80$ & $60.55$ \\
& $d_\textrm{max}=3$ & $61.53$ & $56.50$ & $59.68$ & $67.91$ & $57.87$ & $57.15$ & $63.30$ & $61.58$ & $60.69$ \\
& $\quad$ w/o Clustering & $59.76$ & $55.14$ & $58.29$ & $66.39$ & $59.52$ & $55.32$ & $60.25$ & $60.88$ & $59.44$ \\
& $\quad$ w/o Inst. Embed. & $58.87$ & $57.19$ & $58.15$ & $65.51$ & $59.23$ & $56.65$ & $61.21$ & $61.06$ & $59.73$ \\
& $\quad$ w/o Diversification & $60.75$ & $57.04$ & $59.96$ & $66.25$ & $57.84$ & $57.00$ & $62.89$ & $59.83$ & $60.20$ \\
\midrule
\multirow{5}*{\rotatebox{90}{Variant}} & TopK (Depth Rank) & $59.51$ & $56.24$ & $58.25$ & $64.82$ & $57.26$ & $54.26$ & $61.00$ & $60.59$ & $58.99$ \\
& TopK (Accuracy Rank) & $61.40$ & $56.46$ & $58.90$ & $66.87$ & $58.95$ & $56.14$ & $64.49$ & $61.04$ & $60.53$ \\
& TopK (Usability Rank) & $62.21$ & $56.11$ & $59.27$ & $67.78$ & $58.72$ & $57.97$ & $63.95$ & $61.85$ & $60.98$ \\
& Verbalized CDT & $62.17$ & $57.04$ & $57.93$ & $65.32$ & $58.54$ & $55.01$ & $61.60$ & $60.98$ & $59.82$ \\
& Wikified CDT & $59.21$ & $54.93$ & $56.77$ & $67.76$ & $57.11$ & $54.28$ & $59.15$ & $57.95$ & $58.40$ \\
\bottomrule
\end{tabular}
}
\vspace{-2mm}
\caption{Ablation study and variant experiments. (Bandori part placed in Table~\ref{tab:full_ablation})}
\vspace{-3mm}
\label{tab:ablation}
\end{table*}

As shown in Table~\ref{tab:main}, across all $8$ Fandom artifacts, CDT(-Lite) achieve the best scores, consistently surpassing Vanilla prompting, fine-tuning, RICL, and ETA. Fine-tuning often underperforms even the Vanilla baseline, which can be attributed to the strict chronological split of training and test sets, as fitting tokens in the first half storyline leads to repeating similar actions rather than learning global behavior pattern. RICL and ETA provide noticeable gains over Vanilla in several settings, confirming the value of grounding on past storylines, but they remain clearly behind CDT(-Lite), which benefit from structured, situation-specific codification rather than purely textual aggregation.

CDT also outperforms the “Human Profile” and “Codified Human Profile” baselines, which use human-written profiles as commonly accepted ground truth. While these human profiles are strong and generally superior to unguided or purely data-hungry approaches, CDT(-Lite) obtain higher scores for every Fandom artifact, demonstrating that data-driven profiling under the codification framework can exceed manually constructed descriptions in both coverage and situational fidelity.

On the Bandori benchmark, we observe an even larger margin: CDT(-Lite) achieve the top performance for all $8$ bands, with CDT-Lite slightly ahead of CDT in most cases. Traditional grounding methods (fine-tuning, RICL, ETA) improve over Vanilla but still lag behind our codified approach, and even the best human-written or codified human profiles are outperformed by CDT(-Lite). These results indicate that 1) codified, situation-aware profiling is particularly beneficial for conversational RP, and 2) the lightweight CDT-Lite architecture retains most of the benefits of CDT, often improving further, while enabling cheaper validation, making data-driven codified profiling practical at scale.

\section{Analyses}

Because of the page limitation, we place hyperparameter analysis in Appendix~\ref{apdx:extended_analysis}, efficiency analysis in Appendix~\ref{apdx:efficiency}, and running cases in Appendix~\ref{apdx:profiling_inference_cases}.

\subsection{Ablation Study}

\paragraph{With abolished statements} This ablation verifies the importance of the statement validation stage by simply appending the abolished statements back to the CDT. Comparing with CDT-Lite in Table~\ref{tab:ablation} shows that blindly accepting LLM hypotheses is consistently worse, highlighting the importance of explicit validation in improving reliability.

\paragraph{Node Depth}
We next ablate the maximum traversal depth $d_{\textrm{max}}$, which controls how far a scene is allowed to descend in the tree. The case $d_{\textrm{max}}=1$ degenerates CDT into a flat, codified profile with only root-level rules. Ablation result shows that RP performance generally improves when deepening the CDT, but it also observes a gradual saturation.

\paragraph{Clustering Mechanism}
To study the role of clustering, we consider two variants: (1) \textbf{w/o Clustering} disables clustering altogether and hypothesize triggers on the entire set of scene–action pairs at each node, forcing the LLM to explain a more heterogeneous mixture of behaviors; (2) \textbf{w/o Inst. Embed.} retains clustering but replace instruction-following action embeddings with simpler representations, reducing the semantic alignment between scenes and actions. Both lead to consistent drops compared to CDT-Lite (Table~\ref{tab:ablation}), indicating that (i) grouping semantically similar instances before rule induction and (ii) using instruction-following embeddings to represent actions are both crucial for discovering clean, reusable triggers.

\begin{figure}
    \centering
    \includegraphics[width=0.88\linewidth]{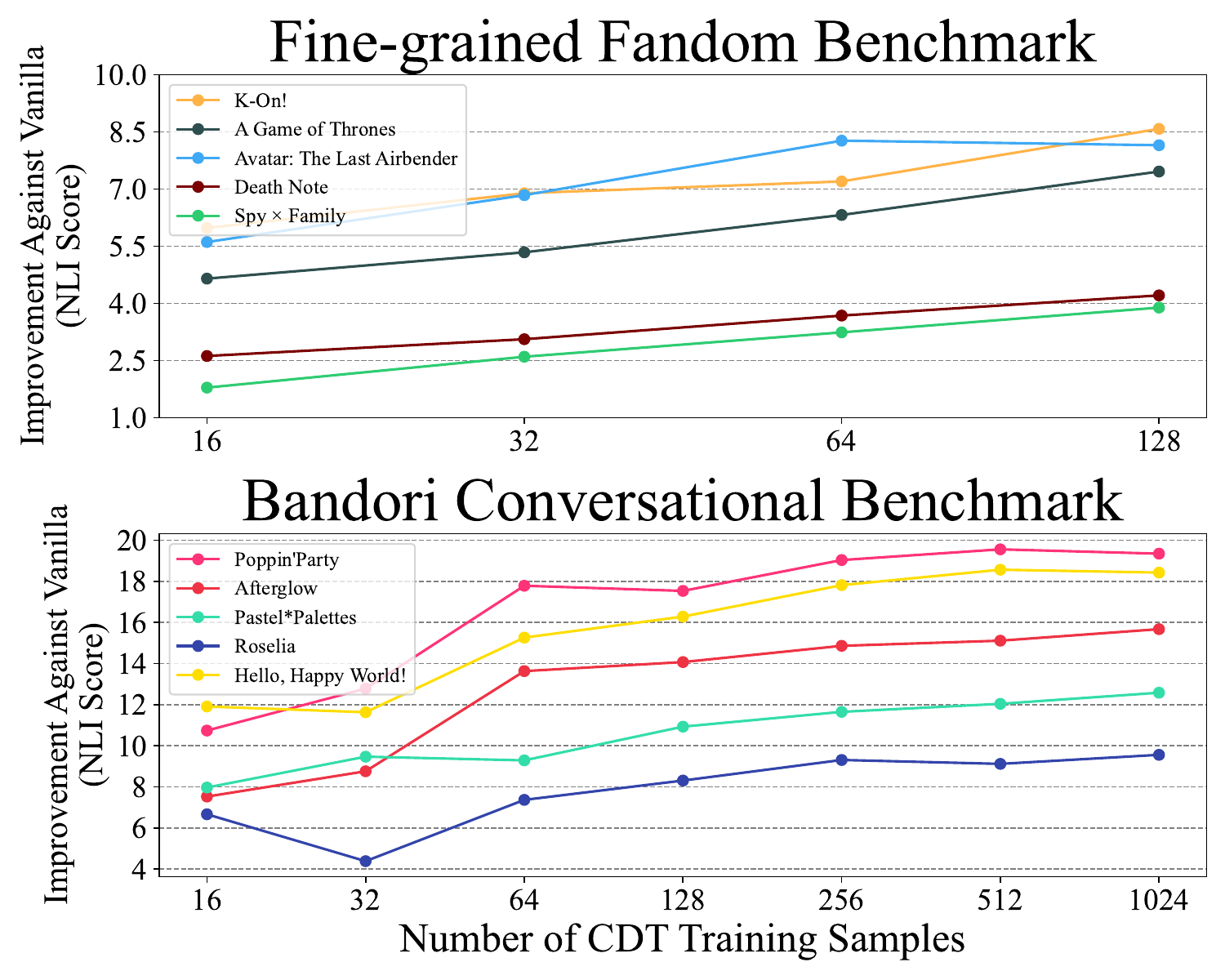}
    \vspace{-3mm}
    \caption{Performance scales up with training data.}
    \label{fig:scaling_law}
    \vspace{-5mm}
\end{figure}

\paragraph{Diversification}
\label{sec:diversification}
As shown in Table~\ref{tab:ablation}, turning off diversification also results in a drop of CDT performance, which validates its contribution in probing more potential candidate behaviors.

\subsection{Variants}

\paragraph{Top-K Policy}
During inference, CDT may activate multiple codified statements along a traversal path, requiring a selection policy. We evaluate three \textbf{Top-K} ranking strategies: \textbf{Depth Rank}, which favors deeper (more specific) nodes; \textbf{Accuracy Rank}, which orders statements by validated accuracy $\frac{r_e}{r_e+r_c}$; and \textbf{Usability Rank}, which prioritizes empirically applicable rules $\frac{r_e}{r_e+r_n+r_c}$. As shown in Table~\ref{tab:ablation}, Usability Rank generally yields the strongest results, suggesting that broadly applicable rules provide better grounding for RP.

\paragraph{CDT-to-Wiki}
We also explore using CDTs as textual profiles. \textbf{Verbalized CDT} linearizes the tree into explicit if-then rules and appends them to the prompt, preserving conditional structure, while \textbf{Wikified CDT} further rewrites these rules into a wiki-style narrative (pipeline described in Appendix~\ref{apdx:profiling_inference_cases}), removing most control-flow cues. Both variants eliminate runtime traversal and can be used by standard RP models. As shown in Table~\ref{tab:ablation}, they remain strong baselines and outperform human-written profiles, despite sacrificing fine-grained situational control and performance.

\subsection{Scaling-up by Data}

We study how CDT scales with additional supervision by varying the number of training scene–action pairs. Figure~\ref{fig:scaling_law} reports CDT performance under different training sizes. For the Fandom benchmark, we select characters with at least $128$ actions and randomly sample $\{16, 32, 64, 128\}$ pairs per character. For the Bandori benchmark, using the richer event-story data, we further scale training up to $1024$ pairs for five bands.

Across both benchmarks, CDT exhibits a clear scaling trend: performance steadily improves as more training data are used. On Fandom, CDT trained with only $64$ pairs already surpasses human-written profile baselines, showing that limited behavioral evidence can outperform manually authored descriptions. On Bandori, gains continue beyond $128$ and up to $1024$ pairs, indicating that RP performance benefits from scaling up the number of interaction data.

\subsection{Relation Modeling}

\begin{table}
\centering
\small
\scalebox{.76}{
\begin{tabular}{llcccc}
\toprule
\multirow{2}*{Character} & \multirow{2}*{Relation with} & \multicolumn{2}{c}{Target Subset} & \multicolumn{2}{c}{Full Test} \\
 &  & CDT & +GD & CDT & +GD \\
\midrule
Suzumiya Haruhi & Kyon & $56.45$ & $\textbf{67.74}$ & $67.91$ & $\textbf{70.14}$ \\
Akiyama Mio & Tainaka Ritsu & $44.93$ & $\textbf{50.00}$ & $52.63$ & $\textbf{54.30}$ \\
Yor Forger & Loid Forger & $64.13$ & $\textbf{70.65}$ & $60.05$ & $\textbf{60.76}$ \\
Alphonse Elric & Edward Elric & $75.00$ & $\textbf{81.82}$ & $63.43$ & $\textbf{64.82}$ \\
Toyama Kasumi & Ichigaya Arisa & $85.95$ & $\textbf{88.02}$ & $85.78$ & $\textbf{86.44}$ \\
Mitake Ran & Aoba Moca & $81.25$ & $\textbf{85.00}$ & $85.12$ & $\textbf{86.68}$ \\
Hikawa Hina & Hikawa Sayo & $85.05$ & $\textbf{88.92}$ & $86.84$ & $\textbf{87.69}$ \\
Hikawa Sayo & Hikawa Hina & $76.52$ & $\textbf{86.87}$ & $76.15$ & $\textbf{80.39}$ \\
Chihaya Anon & Takamatsu Tomori & $69.33$ & $\textbf{72.00}$ & $82.18$ & $\textbf{83.06}$ \\
\bottomrule
\end{tabular}
}
\caption{Performance of Goal-driven (GD) CDT.}
\vspace{-3mm}
\label{tab:goal_driven_cdt}
\end{table}

Beyond general profiling, CDT also supports \emph{goal-driven} specialization toward specific behavioral aspects by \textbf{instructing the trigger-hypothesis prompt}. We demonstrate this capability with \textbf{relation modeling}, which aims to capture characteristic interaction patterns between pairs of characters. For each target character, we filter the training data to retain only actions that occur immediately after a designated related character speaks or acts, and train a relation-specific \emph{goal-driven} CDT that is instructed to focus on their special interaction patterns. At inference time, scenes are routed through this relation-specific CDT whenever the latest context action is taken by the related character; otherwise, the general CDT is used. As shown in Table~\ref{tab:goal_driven_cdt}, this +GD variant consistently outperforms the base CDT on the target subset (cases when interacting with the related character) across all nine character pairs and thus results in gains on the full test set, demonstrating that CDT can model fine-grained relational behaviors without sacrificing generality.

\subsection{Case Study}

Figure~\ref{fig:cdt_wiki} contrasts a human-written character profile with a Wikified CDT for \textit{``Haruhi Suzumiya''}, highlighting differences in granularity and behavioral coverage. While the human profile offers a compact, intuitive summary of key traits, the Wikified CDT expands these traits into systematically grounded patterns that capture a wider range of recurring behaviors and situational triggers. This comparison illustrates how CDT provides more comprehensive behavioral grounding than manually authored profiles.
\section{Conclusion and Future Work}

We present \emph{Codified Decision Trees} (CDT), a data-driven framework that induces executable, situation-specific character profiles from storylines. We show that CDT consistently outperforms previous methods, even using human-written profiles. As future work, we plan to extend CDT to multi-character joint profiling and interaction modeling, continual and online updates from interactive RP logs, and multi-modal settings where scenes include visual or game-state signals, moving toward more broadly controllable, behavior-grounded agents beyond the RP domain.
\section*{Limitations}

While CDT demonstrates strong performance and interpretability, it currently relies only on storyline-derived scene-action pairs and does not incorporate pre-designed persona information (e.g., canonical traits or author intent). We adopt this setting to isolate the effect of data-driven profiling and to avoid introducing external human priors that may be inconsistent or unavailable across domains; incorporating such information as soft priors for trigger induction is a natural extension. CDT is also constructed offline and remains static, which simplifies validation and benchmarking but does not capture evolving narratives; enabling incremental or continual CDT updates as new storylines unfold is an important direction for future work. Finally, although we focus on role-playing, the CDT framework is general and could be extended to other situation-aware behavior modeling tasks, such as task-oriented agents or embodied decision-making.

\bibliography{custom}

\clearpage

\appendix

\begin{table*}[ht]
\centering
\small
\scalebox{.99}{
\begin{tabular}{lcccccccccc}
\toprule
Fandom & {Haruhi} & {K-On!} & {S$\times$F} & {DN} & {FMA} & {JOJO} & {AGOT} & {ATLA} \\
\midrule
\#Main Character & 5 & 5 & 3 & 5 & 5 & 7 & 11 & 4 \\
\#Episode & 28 & 57 & 116 & 108 & 108 & 152 & 73 & 61 \\
\#Action & 991 & 2555 & 7688 & 5006 & 3349 & 2958 & 12073 & 8619 \\
\#Action$_{\textrm{Main Character}}$ & 781 & 1882 & 3341 & 2738 & 1351 & 1578 & 4859 & 4248 \\
\#Avg. Action Length & 12.15 & 10.51 & 11.73 & 13.13 & 12.58 & 11.81 & 12.42 & 11.58 \\
\midrule
\midrule
Bandori & {PoPiPa} & {AG} & {PasuPare} & {Roselia} & {HHW} & {Monica} & {RAS} & {MyGO} \\
\midrule
\#Main Character & 5 & 5 & 5 & 5 & 5 & 5 & 5 & 5 \\
\#Episode & 20 & 20 & 20 & 20 & 20 & 20 & 25 & 41 \\
\#Action & 1226 & 1053 & 968 & 1079 & 1122 & 1040 & 1183 & 2050 \\
\#Action$_{\textrm{Main Character}}$ & 1080 & 914 & 791 & 873 & 827 & 966 & 795 & 1620 \\
\#Avg. Action Length & 13.18 & 16.61 & 21.01 & 21.95 & 21.60 & 19.66 & 15.64 & 16.48 \\
\midrule
\midrule
Bandori (Events) & {PoPiPa} & {AG} & {PasuPare} & {Roselia} & {HHW} & {Monica} & {RAS} & {MyGO} \\
\midrule
\#Main Character & 5 & 5 & 5 & 5 & 5 & 5 & 5 & 5 \\
\#Episode & \multicolumn{8}{c}{1498} \\
\#Action & \multicolumn{8}{c}{77182} \\
\#Action$_{\textrm{Main Character}}$ & 12553 & 11365 & 12058 & 11821 & 10287 & 4758 & 2863 & 745 \\
\#Avg. Action Length & 19.63 & 19.87 & 21.61 & 21.07 & 20.82 & 20.73 & 19.53 & 15.90 \\
\bottomrule
\end{tabular}
}
\vspace{-2mm}
\caption{Statistics of benchmarks (\textit{Fine-grained Fandom Benchmark} and \textit{Bandori Conversational Benchmark}) used in the experiments.}
\vspace{-3mm}
\label{tab:stats_benchmark}
\end{table*}

\begin{table*}[ht]
\centering
\small
\scalebox{.99}{
\begin{tabular}{lcccccccccc}
\toprule
Fandom & {Haruhi} & {K-On!} & {S$\times$F} & {DN} & {FMA} & {JOJO} & {AGOT} & {ATLA} \\
\midrule
\#Node & 3.80 & 32.80 & 162.33 & 45.80 & 73.40 & 32.14 & 79.73 & 309.50 \\
\#Statement & 16.80 & 90.80 & 259.67 & 135.20 & 88.20 & 50.14 & 159.55 & 497.25 \\
\#Avg. Statement Length & 18.03 & 19.75 & 20.46 & 19.63 & 19.46 & 18.49 & 20.56 & 20.20 \\
\midrule
\midrule
Bandori & {PoPiPa} & {AG} & {PasuPare} & {Roselia} & {HHW} & {Monica} & {RAS} & {MyGO} \\
\midrule
\#Node & 10.40 & 28.40 & 41.40 & 21.80 & 5.20 & 52.00 & 59.00 & 9.80 \\
\#Statement & 61.00 & 107.40 & 99.40 & 68.20 & 27.40 & 112.00 & 114.60 & 82.40 \\
\#Avg. Statement Length & 18.35 & 19.23 & 19.74 & 20.62 & 18.23 & 18.37 & 19.05 & 18.66 \\
\bottomrule
\end{tabular}
}
\vspace{-2mm}
\caption{Statistics of CDTs used in the experiments.}
\vspace{-3mm}
\label{tab:stats_cdt}
\end{table*}

\begin{table*}
\centering
\small
\scalebox{.9}{
\begin{tabular}{llccccccccc}
\toprule
\multicolumn{2}{l}{\textbf{Bandori}} & {PoPiPa} & {AG} & {PasuPare} & {Roselia} & {HHW} & {Monica} & {RAS} & {MyGO} & {Avg.} \\
\midrule
\multicolumn{2}{l}{CDT-Lite} & $88.38$ & $80.49$ & $82.47$ & $72.81$ & $79.66$ & $78.67$ & $79.51$ & $70.33$ & $79.04$ \\
\midrule
\multirow{7}*{\rotatebox{90}{Ablation}}& $\quad$ w/ Abolished Statements & $86.90$ & $76.35$ & $78.69$ & $73.52$ & $78.52$ & $77.44$ & $79.54$ & $67.90$ & $77.36$ \\
& $d_\textrm{max}=1$ & $87.15$ & $77.23$ & $82.19$ & $74.07$ & $77.88$ & $77.96$ & $76.97$ & $69.77$ & $77.90$ \\
& $d_\textrm{max}=2$ & $87.40$ & $79.79$ & $82.85$ & $72.71$ & $79.49$ & $78.44$ & $79.40$ & $70.04$ & $78.77$ \\
& $d_\textrm{max}=3$ & $88.24$ & $79.84$ & $81.76$ & $72.90$ & $79.63$ & $78.97$ & $79.06$ & $70.75$ & $78.89$ \\
& $\quad$ w/o Clustering & $86.99$ & $79.38$ & $80.67$ & $73.22$ & $80.52$ & $78.21$ & $77.43$ & $69.70$ & $78.27$ \\
& $\quad$ w/o Inst. Embed. & $86.97$ & $77.66$ & $78.78$ & $75.91$ & $78.16$ & $79.16$ & $78.93$ & $67.00$ & $77.82$ \\
& $\quad$ w/o Diversification & $88.74$ & $80.24$ & $80.97$ & $73.38$ & $79.13$ & $78.26$ & $78.10$ & $69.97$ & $78.60$ \\
\midrule
\multirow{5}*{\rotatebox{90}{Variant}} & TopK (Depth Rank) & $87.58$ & $76.78$ & $79.57$ & $71.77$ & $78.93$ & $75.77$ & $82.24$ & $68.86$ & $77.69$ \\
& TopK (Accuracy Rank) & $87.73$ & $81.26$ & $81.62$ & $72.65$ & $79.82$ & $76.19$ & $80.55$ & $69.85$ & $78.71$ \\
& TopK (Usability Rank) & $88.17$ & $81.28$ & $83.76$ & $75.19$ & $78.92$ & $78.84$ & $77.72$ & $70.90$ & $79.35$ \\
& Verbalized CDT & $88.57$ & $79.92$ & $81.32$ & $75.18$ & $78.39$ & $78.01$ & $78.10$ & $70.12$ & $78.70$ \\
& Wikified CDT & $86.56$ & $75.92$ & $82.56$ & $73.24$ & $81.21$ & $77.14$ & $78.20$ & $64.35$ & $77.40$ \\
\bottomrule
\end{tabular}
}
\vspace{-2mm}
\caption{Bandori part of ablation study and variant experiments.}
\vspace{-3mm}
\label{tab:full_ablation}
\end{table*}

\begin{table*}
\centering
\small
\scalebox{.99}{
\begin{tabular}{lcccccccccc}
\toprule
Fandom & {Haruhi} & {K-On!} & {S$\times$F} & {DN} & {FMA} & {JOJO} & {AGOT} & {ATLA} \\
\midrule
Vanilla & $43.60$ & $44.20$ & $47.16$ & $51.47$ & $43.71$ & $41.45$ & $42.56$ & $43.18$ \\
Human Profile & $42.70$ & $49.31$ & $53.98$ & $54.26$ & $47.97$ & $49.42$ & $49.14$ & $45.13$ \\
CDT-Lite (Ours) & $54.30$ & $53.88$ & $58.61$ & $60.07$ & $49.82$ & $52.49$ & $50.92$ & $51.92$ \\
\midrule
Vanilla & $50.47$ & $49.73$ & $51.67$ & $56.97$ & $55.74$ & $49.57$ & $49.46$ & $52.77$ \\
Human Profile & $50.58$ & $52.87$ & $55.31$ & $60.45$ & $53.12$ & $50.96$ & $52.38$ & $53.12$ \\
CDT-Lite (Ours) & $54.75$ & $53.75$ & $56.00$ & $62.57$ & $53.16$ & $49.74$ & $53.14$ & $54.45$ \\
\midrule
Human Profile (8B) & $55.87$ & $55.86$ & $59.14$ & $64.75$ & $58.54$ & $55.11$ & $59.35$ & $57.98$ \\
\midrule
\midrule
Bandori & {PoPiPa} & {AG} & {PasuPare} & {Roselia} & {HHW} & {Monica} & {RAS} & {MyGO} \\
\midrule
Vanilla & $58.38$ & $56.44$ & $60.14$ & $56.57$ & $58.50$ & $56.06$ & $59.97$ & $53.60$ \\
Human Profile & $65.34$ & $62.90$ & $66.75$ & $61.03$ & $65.76$ & $55.56$ & $64.73$ & $54.75$ \\
CDT-Lite (Ours) & $78.58$ & $80.05$ & $77.47$ & $68.14$ & $78.90$ & $73.42$ & $73.68$ & $61.16$ \\
\midrule
Vanilla & $63.91$ & $62.27$ & $66.49$ & $60.90$ & $64.63$ & $63.06$ & $62.95$ & $59.62$ \\
Human Profile & $71.96$ & $68.13$ & $74.00$ & $67.64$ & $70.64$ & $65.16$ & $69.27$ & $59.39$ \\
CDT-Lite (Ours) & $79.13$ & $77.72$ & $77.29$ & $67.46$ & $79.26$ & $74.18$ & $74.30$ & $66.51$ \\
\midrule
Human Profile (8B) & $73.73$ & $72.43$ & $77.11$ & $70.08$ & $73.14$ & $68.08$ & $71.74$ & $63.91$ \\
\bottomrule
\end{tabular}
}
\vspace{-2mm}
\caption{RP performance on variant RP models (1B and 3B), compared with key baselines.}
\vspace{-3mm}
\label{tab:rp_model_var}
\end{table*}

\section{Statistics Information}
\label{apdx:stats}

Tables~\ref{tab:stats_benchmark} and~\ref{tab:stats_cdt} summarize the statistics of the benchmarks and the resulting CDTs used in our experiments. The fine-grained Fandom benchmark spans eight artifacts with diverse narrative lengths and cast sizes, ranging from relatively compact series (e.g., \textit{Haruhi}) to long-running storylines with large ensembles (e.g., \textit{AGOT} and \textit{ATLA}). The Bandori conversational benchmark covers all eight bands with balanced main-character counts, while the event-story extension substantially increases data scale, yielding over $77$K actions in total. Across datasets, average action lengths remain stable, indicating consistent annotation granularity.

Table~\ref{tab:stats_cdt} reports structural statistics of the induced CDTs. Despite large variation in data volume, the resulting trees remain shallow and compact, with moderate numbers of nodes and statements per character, reflecting the effectiveness of validation and pruning. Average statement lengths are consistent across artifacts and benchmarks, suggesting that CDT induces comparable, interpretable rule granularity in different narrative domains. These statistics further support that CDT scales to long storylines without uncontrolled growth in profile complexity.

\section{Setup Details}
\label{apdx:hyperparameter}

In our experiments, we set $\theta_{\textrm{accept}}=0.75$, $\theta_{\textrm{reject}}=0.50$, $\theta_{\textrm{f}}=0.75$, max depth $=4$, $|\mathcal{D}|_{min}$ for node growing $=16$, $3$ hypotheses per cluster, minimal $16$ cases per cluster, and a maximum of $8$ clusters per node for all main experiments. For distillation in CDT-Lite, we train the \texttt{deberta-v3-base} discriminator with AdamW~\citep{adamw} initialized by $10^{-5}$ learning rate, $32$ batch size for $1$ epoch. 

For baselines, \textbf{Fine-tuning} adopts LoRA~\citep{lora} fine-tuning with batch size $16$, AdamW initialized by $2\times 10^{-4}$, which enables smooth loss drop. We report the result after the first epoch since the performance continues dropping while fitting on the first half storyline. \textbf{RICL} retrieves the scene with the $8$ most similar scene based on \texttt{qwen3-embedding-8b}. \textbf{ETA} segment scene-action pairs by $16$ samples in each block, with prompts available in Appendix~\ref{fig:baseline_prompts}. \textbf{Human profile and codified human profile}'s setups follow the previous codification work~\citep{codification} using the same distilled discriminator for CDT-Lite.

\section{More Variant Studies}
\label{apdx:extended_analysis}

The full ablation and variant study results are placed in Table~\ref{tab:full_ablation}, which reaches a similar conclusion about component contribution, top statement selection strategy, and verbalization effect.

\begin{table*}
\centering
\small
\scalebox{.99}{
\begin{tabular}{lcccccccccc}
\toprule
Fandom & {Haruhi} & {K-On!} & {S$\times$F} & {DN} & {FMA} & {JOJO} & {AGOT} & {ATLA} \\
\midrule
\texttt{gpt-4.1-mini} & $60.45$ & $58.66$ & $59.07$ & $67.27$ & $59.19$ & $55.35$ & $62.06$ & $60.95$ \\
\texttt{gpt-4.1} & $62.17$ & $57.24$ & $59.79$ & $67.00$ & $59.04$ & $57.26$ & $64.27$ & $61.32$ \\
\texttt{qwen3-coder} & $61.37$ & $57.05$ & $58.83$ & $66.39$ & $56.60$ & $56.88$ & $61.07$ & $60.14$ \\
\midrule
\midrule
Bandori & {PoPiPa} & {AG} & {PasuPare} & {Roselia} & {HHW} & {Monica} & {RAS} & {MyGO} \\
\midrule
\texttt{gpt-4.1-mini} & $85.62$ & $80.24$ & $79.26$ & $73.66$ & $79.63$ & $77.80$ & $77.49$ & $73.10$ \\
\texttt{gpt-4.1} & $88.38$ & $80.49$ & $82.47$ & $72.81$ & $79.66$ & $78.67$ & $79.51$ & $70.33$ \\
\texttt{qwen3-coder} & $82.95$ & $81.13$ & $79.83$ & $74.06$ & $80.46$ & $77.37$ & $78.57$ & $71.09$ \\
\bottomrule
\end{tabular}
}
\vspace{-2mm}
\caption{RP performance on variant codifiers, compared with key baselines.}
\vspace{-3mm}
\label{tab:codifier_var}
\end{table*}

\paragraph{RP Model Variants}
Table~\ref{tab:rp_model_var} reports results when varying the RP backbone while keeping CDT-Lite unchanged. We evaluate smaller RP models (1B and 3B) and compare CDT-Lite against Vanilla prompting and human-written Human Profiles. Across both the Fandom and Bandori benchmarks, CDT-Lite consistently delivers large gains over Vanilla and Human Profile for all model sizes, demonstrating that codified, situation-aware grounding is effective even with limited-capacity RP models. While absolute performance improves with stronger RP backbones, the relative advantage of CDT-Lite remains stable, indicating that its benefits are largely orthogonal to model scale. Notably, CDT-Lite with smaller RP models can approach or even surpass the performance of larger models grounded with human-written profiles, highlighting its practical value in resource-constrained settings.

\paragraph{Codifier Variants}
Table~\ref{tab:codifier_var} compares different codifiers used for trigger hypothesis and validation when constructing CDT-Lite. Using \texttt{gpt-4.1} as the codifier generally yields the strongest performance, while \texttt{gpt-4.1-mini} or \texttt{qwen3-coder} achieves slightly lower but still competitive results on both benchmarks. The relatively small gap between these settings suggests that CDT-Lite is robust to codifier strength and can be instantiated with lighter or open-source models at reduced cost. These results further support the scalability of CDT, showing that high-quality codified profiles can be induced without relying exclusively on the strongest available LLMs.

\begin{table}[H]
\centering
\small
\scalebox{.9}{
\begin{tabular}{llcccc}
\toprule
{$\theta_\textrm{accept}$} &  & $0.70$ & $0.75$ & $0.80$ & $0.85$ \\
\midrule
\multirow{3}*{\rotatebox{90}{$\theta_\textrm{reject}$}} & $0.40$ & $69.28$ & $69.74$ & $69.35$ & $68.82$ \\
 & $0.50$ & $69.13$ & $70.16$ & $70.53$ & $69.54$ \\
 & $0.60$ & $66.05$ & $67.82$ & $69.33$ & $68.89$ \\
\midrule
\multicolumn{2}{l}{\#Hypothesis} & $1$ & $2$ & $3$ & $4$ \\
\midrule
 &  & $68.53$ & $69.46$ & $70.16$ & $70.33$ \\
\midrule
\multicolumn{2}{l}{\#Max Cluster} & $4$ & $8$ & $12$ & $16$ \\
\midrule
 &  & $68.92$ & $70.16$ & $70.22$ & $70.43$ \\
\bottomrule
\end{tabular}
}
\vspace{-2mm}
\caption{CDT performance with different hyperparameter setups.}
\vspace{-3mm}
\label{tab:hyperparameter}
\end{table}

\paragraph{Hyperparameter Setup}
Table~\ref{tab:hyperparameter} reports the sensitivity of CDT performance to different hyperparameter configurations, evaluated on a sampled $10\%$ subset of the test set. We vary the trigger acceptance threshold $\theta_{\textrm{accept}}$, rejection threshold $\theta_{\textrm{reject}}$, the number of hypotheses proposed per cluster, and the maximum number of clusters expanded at each node. Overall, performance is relatively stable across a wide range of settings, indicating that CDT is not overly sensitive to precise hyperparameter tuning. We apply $\theta_{\textrm{accept}}=0.75$, $\theta_{\textrm{reject}}=0.50$, $3$ hypotheses per cluster, and a maximum of $8$ clusters per node for all main experiments, which achieves a strong balance between performance and computational cost.

\clearpage

\section{Evaluation Generality Validation}
\label{apdx:extended_metric}

\begin{figure}[ht]
    \centering
    \includegraphics[width=0.9\linewidth]{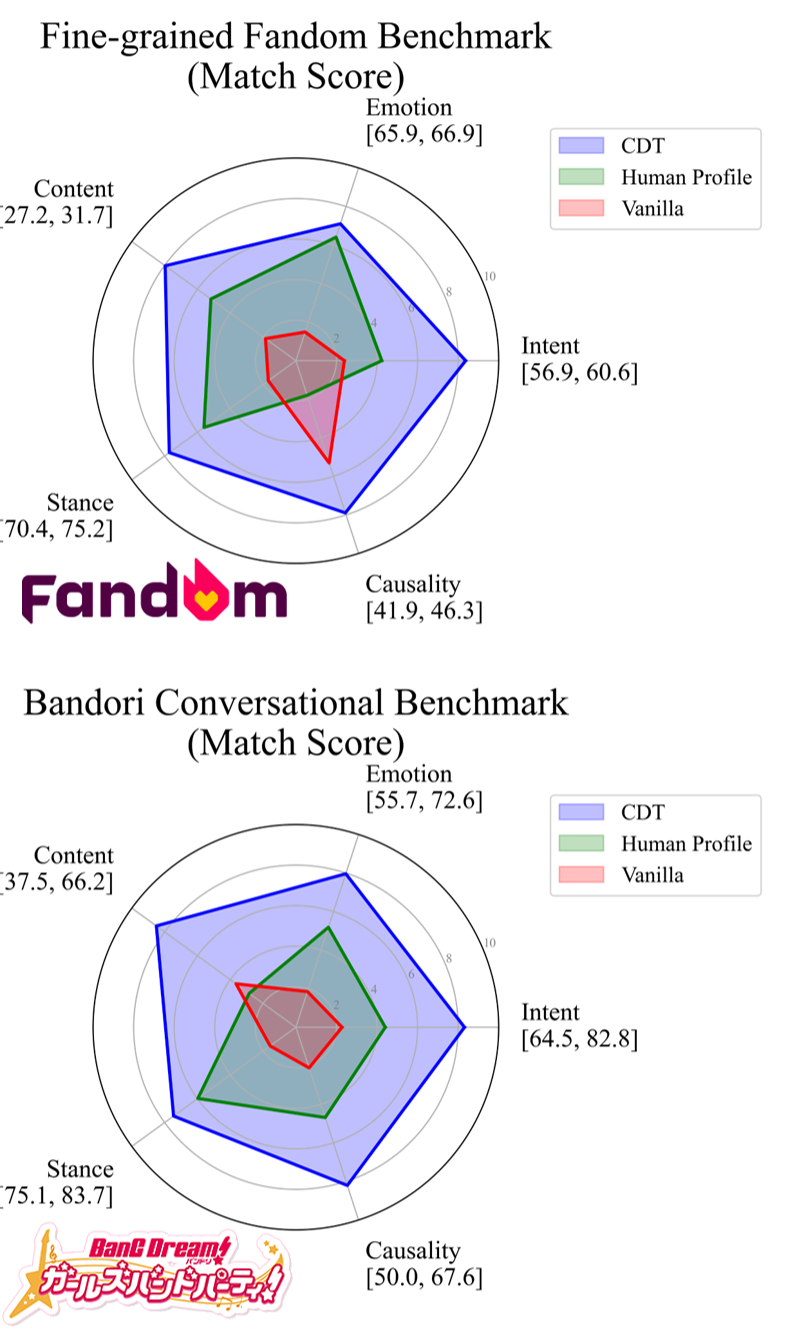}
    \caption{The comparison between grounding methods with matching scores. (CDT represents CDT-Lite)}
    \label{fig:match_score}
\end{figure}

\paragraph{Cross-metric and Manual Validation} Figure~\ref{fig:match_score} further shows that the NLI-based matching score is consistent with prior LLM-based multi-dimensional judgment protocols, validating its use as a reliable automatic evaluator. Across both benchmarks, CDT exhibits clear and systematic gains on all evaluated dimensions, \emph{intent}, \emph{emotion}, \emph{content}, \emph{stance}, and \emph{causality}, with particularly strong improvements on intent and causality, reflecting CDT’s ability to condition behavior on situation-specific triggers rather than generic traits. Improvements in content and stance indicate that CDT grounding helps the RP model select contextually appropriate details and attitudes, while emotion gains suggest better alignment with character affect under varying scenes. We also verify that the trends observed in NLI scores closely match those of manual inspections: based on $200$ samples for each class, the NLI discrimination achieves $90.50\%$, $92.00\%$, and $88.50\%$ consistency with human judgment on \textit{``entailed''}, \textit{``neutral''}, and \textit{``contradicted''} while matching score has $89.00\%$ and $90.50\%$ consistency on \textit{``matched''} and \textit{``mismatched''}. This confirms that both NLI accuracy and the aggregated matching score serve as dependable proxies for RP quality.

\begin{figure}[ht]
    \centering
    \includegraphics[width=0.99\linewidth]{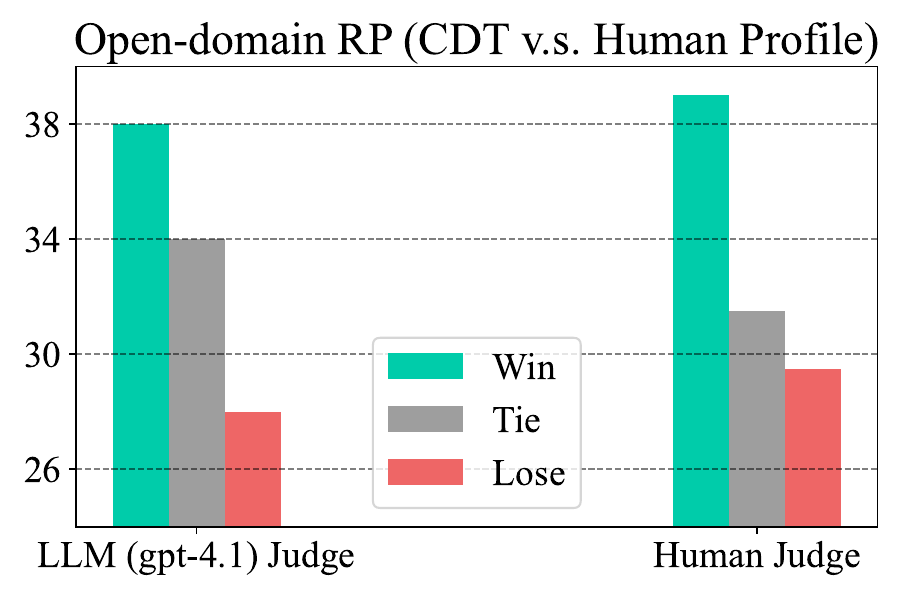}
    \caption{Comparison on open-ended RP by Human and LLM judgment. (CDT represents CDT-Lite)}
    \label{fig:cdt_preference}
\end{figure}

\paragraph{Open-domain Test}
We further include two types of open-domain evaluations to assess whether the induced CDTs generalize beyond the storylines on which they are trained. On the Fandom benchmark, we propose $200$ starting scenes that initiate novel interactive settings (e.g., battles, casual conversations) between characters from the same artifact, and roll out $10$ turns of interaction for each scene. CDT is compared against the Human Profile baseline, with responses evaluated by both an LLM judge (\texttt{gpt-4.1}) and human annotators. As shown in Figure~\ref{fig:cdt_preference}, CDT is preferred over Human Profile in $38.0\%$ versus $26.0\%$ of cases by the LLM judge, and $39.0\%$ versus $27.5\%$ by human evaluators, indicating that CDT generalizes better to unseen interactive scenarios.

On the Bandori benchmark, we sample $100$ scene–action pairs per band member from the wild event stories (drawn from a pool of $77.2$K actions) and evaluate CDTs trained only on the first half of the main band stories. Table~\ref{tab:codifier_var} shows that CDT consistently outperforms Vanilla prompting and Human Profiles across all eight bands, demonstrating robust out-of-domain generalization from curated storylines to heterogeneous, real-world conversational data. Together, these results confirm that codified, situation-aware profiles learned by CDT transfer effectively to novel scenes and interaction patterns beyond the original training distributions.

\begin{table}[H]
\centering
\small
\scalebox{.95}{
\begin{tabular}{lcccccccccc}
\toprule
Bandori & {PoPiPa} & {AG} & {PasuPare} & {Roselia} \\
\midrule
Vanilla & $71.80$ & $70.20$ & $68.60$ & $70.60$ \\
Human Profile & $74.60$ & $69.90$ & $73.70$ & $73.50$ \\
CDT & $85.40$ & $81.40$ & $85.10$ & $74.50$ \\
\midrule
Bandori & {HHW} & {Monica} & {RAS} & {MyGO} \\
\midrule
Vanilla & $72.70$ & $69.60$ & $72.20$ & $63.10$ \\
Human Profile & $79.00$ & $70.00$ & $74.90$ & $69.80$ \\
CDT & $80.60$ & $81.10$ & $81.10$ & $72.00$ \\
\bottomrule
\end{tabular}
}
\vspace{-2mm}
\caption{Out-of-domain (OOD) RP performance evaluation with CDT trained on main band stories tested on wild event stories.}
\vspace{-3mm}
\label{tab:codifier_var}
\end{table}

\section{Cost and Efficiency}
\label{apdx:efficiency}

\paragraph{Boosted CDT}
To further improve efficiency, we introduce \textbf{Boosted CDT}, which retains only the most important hypothesized triggers (eight in our implementation) before validation. This reduces redundancy and focuses validation on the core behavioral logic, at the cost of omitting some fine-grained details. As shown in Table~\ref{tab:boosted_cdt}, Boosted CDT achieves performance close to CDT-Lite on both Fandom and Bandori, while reducing the number of nodes in the tree by an order of magnitude.

\paragraph{Training and Grounding Efficiency}
Table~\ref{tab:efficiency} summarizes the computational efficiency of CDT-Lite and Boosted CDT. A key observation is that CDT construction is \emph{validation-heavy}: the number of discrimination calls far exceeds the number of generation calls, since each hypothesized trigger must be tested against all scene–action pairs. This motivates the use of distilled discriminators (e.g., \texttt{deberta-v3-base}) to replace expensive LLM validators. With distilled discrimination, CDT becomes practical to train at scale, as validation, which is the dominant cost, can be performed cheaply without significant performance degradation. 

The number of nodes in each CDT (Table~\ref{tab:boosted_cdt}, bottom rows) also approximates the maximal traversal steps needed during inference. Because traversal is performed via distilled discriminators rather than full LLM calls, its runtime cost is negligible compared to generating the final RP response. Consequently, CDT grounding remains lightweight at inference time, and Boosted CDT offers additional speedups in trade-off for lower accuracy by further shrinking the tree.

\begin{table*}
\centering
\small
\scalebox{.99}{
\begin{tabular}{llcccccccccc}
\toprule
\multicolumn{2}{l}{Fandom} & {Haruhi} & {K-On!} & {S$\times$F} & {DN} & {FMA} & {JOJO} & {AGOT} & {ATLA} \\
\midrule
\multirow{3}*{NLI} & Human Profile & $55.87$ & $55.86$ & $59.14$ & $64.75$ & $58.54$ & $55.11$ & $59.35$ & $57.98$ \\
& CDT-Lite & $62.17$ & $57.24$ & $59.79$ & $67.00$ & $59.04$ & $57.26$ & $64.27$ & $61.32$ \\
& Boosted CDT-Lite & $61.70$ & $57.19$ & $60.04$ & $66.29$ & $58.53$ & $58.15$ & $61.42$ & $60.00$ \\
\midrule
\multirow{2}*{\#Node} & CDT-Lite & $3.80$ & $32.80$ & $162.33$ & $45.80$ & $73.40$ & $32.14$ & $79.73$ & $309.50$ \\
& Boosted CDT-Lite & $1.60$ & $6.60$ & $8.33$ & $2.40$ & $5.60$ & $4.71$ & $4.91$ & $10.00$ \\
\midrule
\midrule
\multicolumn{2}{l}{Bandori} & {PoPiPa} & {AG} & {PasuPare} & {Roselia} & {HHW} & {Monica} & {RAS} & {MyGO} \\
\midrule
\multirow{3}*{NLI} & Human Profile & $73.73$ & $72.43$ & $77.11$ & $70.08$ & $73.14$ & $68.08$ & $71.74$ & $63.91$ \\
& CDT-Lite & $88.38$ & $80.49$ & $82.47$ & $72.81$ & $79.66$ & $78.67$ & $79.51$ & $70.33$ \\
& Boosted CDT-Lite & $86.96$ & $76.19$ & $79.81$ & $74.88$ & $78.34$ & $75.44$ & $78.29$ & $69.74$ \\
\midrule
\multirow{2}*{\#Node} & CDT-Lite & $10.40$ & $28.40$ & $41.40$ & $21.80$ & $5.20$ & $52.00$ & $59.00$ & $9.80$ \\
& Boosted CDT-Lite & $5.80$ & $6.60$ & $9.80$ & $4.60$ & $3.20$ & $4.40$ & $10.40$ & $1.80$ \\
\bottomrule
\end{tabular}
}
\vspace{-2mm}
\caption{Further boosting the efficiency of CDT and resulted performance.}
\vspace{-3mm}
\label{tab:boosted_cdt}
\end{table*}

\begin{table}[H]
\centering
\small
\scalebox{.99}{
\begin{tabular}{llcccc}
\toprule
 & Metric & CDT-Lite & Boosted CDT-Lite \\
\midrule
\multirow{3}*{\rotatebox{90}{Fandom}} & \#Gen. Call & $58.47$ & $23.28$ \\
 & \#Disc. Call & $40.62$K & $9.04$K \\
 & \#Node & $80.13$ & $5.18$ \\
 \midrule
\multirow{3}*{\rotatebox{90}{Bandori}}  & \#Gen. Call & $37.35$ & $18.57$ \\
 & \#Disc. Call & $12.06$K & $3.08$K \\
 & \#Node & $28.50$ & $5.82$ \\
\bottomrule
\end{tabular}
}
\vspace{-2mm}
\caption{Training and Grounding efficiency of (Boosted) CDT-Lite per character.}
\vspace{-3mm}
\label{tab:efficiency}
\end{table}

\clearpage

\section{Profiling and Inference Cases}
\label{apdx:profiling_inference_cases}

\begin{figure*}
    \centering
    \includegraphics[width=0.9\linewidth]{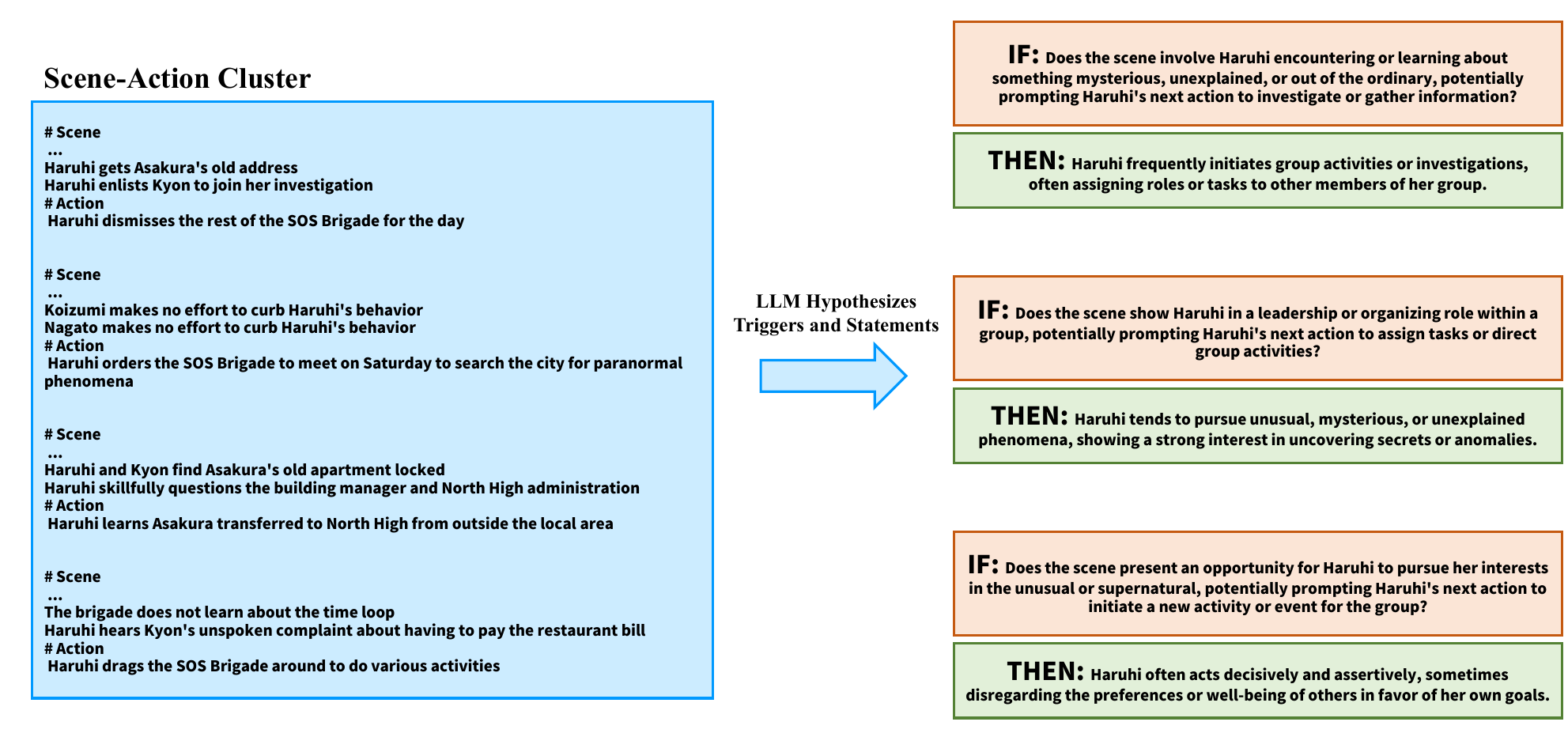}
    \vspace{-3mm}
    \caption{A running case for behavior hypothesis in the CDT building.}
    \label{fig:running_hypothesis}
    \vspace{-5mm}
\end{figure*}

\begin{figure*}
    \centering
    \includegraphics[width=0.9\linewidth]{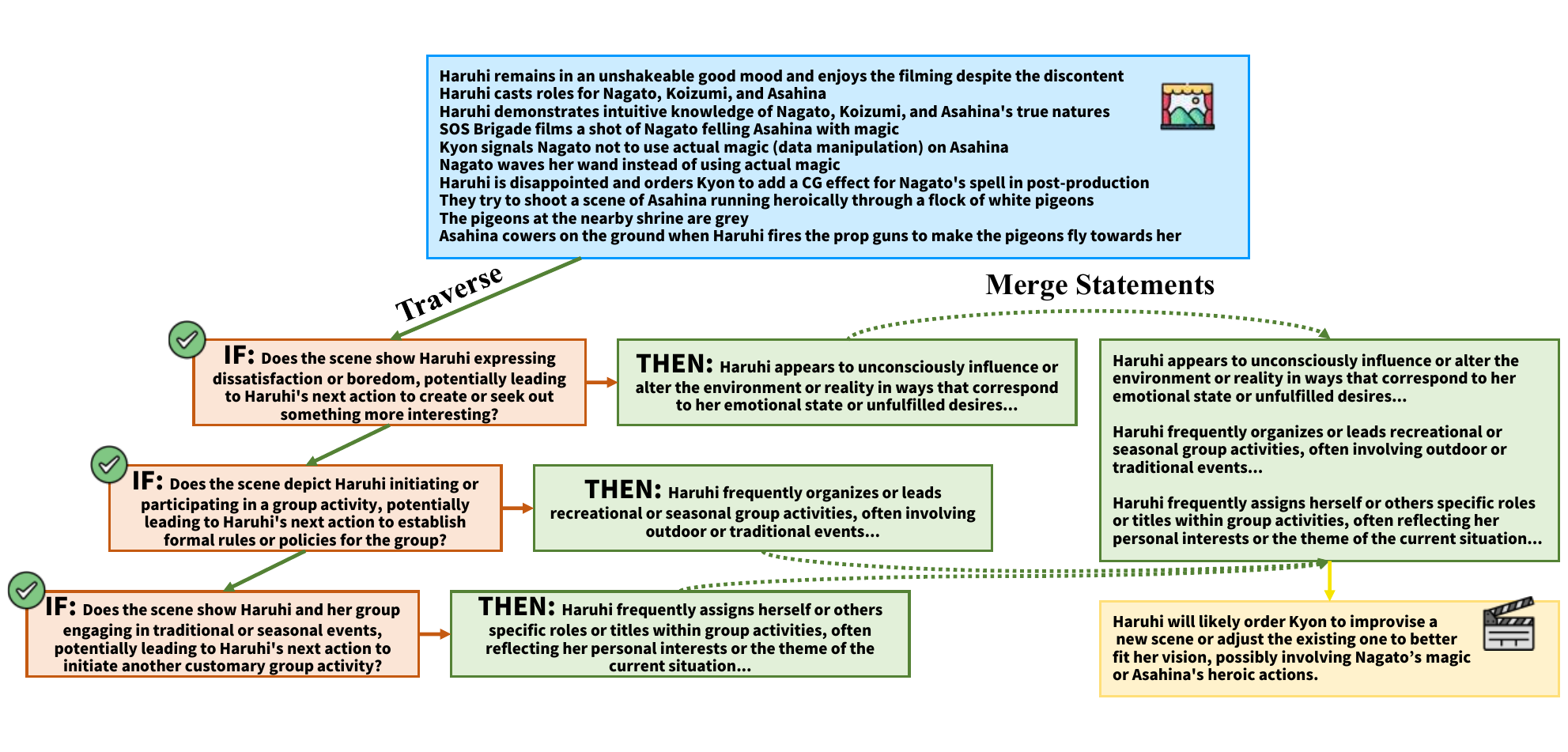}
    \vspace{-3mm}
    \caption{A running case for tree traversal in the CDT grounding.}
    \label{fig:running_inference}
    \vspace{-5mm}
\end{figure*}

We visualize running cases of CDT behavior hypothesis construction in Figure~\ref{fig:running_hypothesis}, and the traversal (grounding) stage in Figure~\ref{fig:running_inference}, using \textit{``Haruhi Suzumiya''} as the instance.

Figure~\ref{fig:running_hypothesis} shows how CDT generalizes from clustered scene--action evidence into reusable \texttt{IF--THEN} hypotheses.
Each cluster groups semantically similar scenes with an observed action, and the LLM summarizes them into trigger conditions (\texttt{IF}) and behavioral statements (\texttt{THEN}).
In this example, the hypotheses characterize Haruhi as someone who eagerly pursues unusual or unexplained phenomena, initiates group activities or investigations, and frequently assigns roles or tasks to other SOS Brigade members.
The extracted rules are stored as nodes in the CDT for later retrieval.

At inference time, Figure~\ref{fig:running_inference} illustrates traversal over the hypothesis tree given a new scene.
The system checks candidate \texttt{IF} conditions, selects those supported by the scene, and collects the corresponding \texttt{THEN} statements along the traversal path.
These statements are merged into a compact guidance set that constrains the response generation.
In the running case, the merged guidance supports predicting Haruhi will actively steer the filming (e.g., instructing Kyon to modify a scene to match her vision), grounded by an entailed reference action from the evidence (e.g., Haruhi using prop guns to drive pigeons toward Asahina).

\paragraph{Wikification pipeline} The CDT is first linearized into a verbalized CDT. Then, the prompts in Figure~\ref{fig:baseline_prompts} are used to propose chapters and fill in chapter contents to build the wikified CDT.

\begin{figure*}
    \centering
    \includegraphics[width=0.9\linewidth]{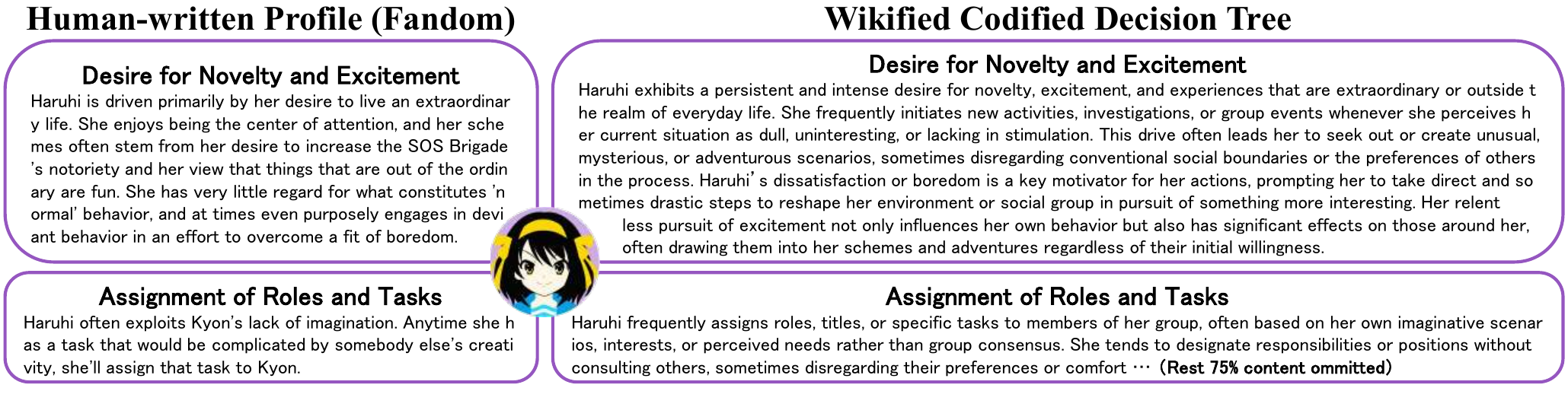}
    \vspace{-3mm}
    \caption{Granularity and coverage comparison between human profile and CDT. (Full wikified CDT in Figure~\ref{fig:full_wikified_cdt_1},~\ref{fig:full_wikified_cdt_2}, and~\ref{fig:full_wikified_cdt_3})}
    \label{fig:cdt_wiki}
    \vspace{-5mm}
\end{figure*}

\begin{figure*}
    \centering
    \includegraphics[width=0.9\linewidth]{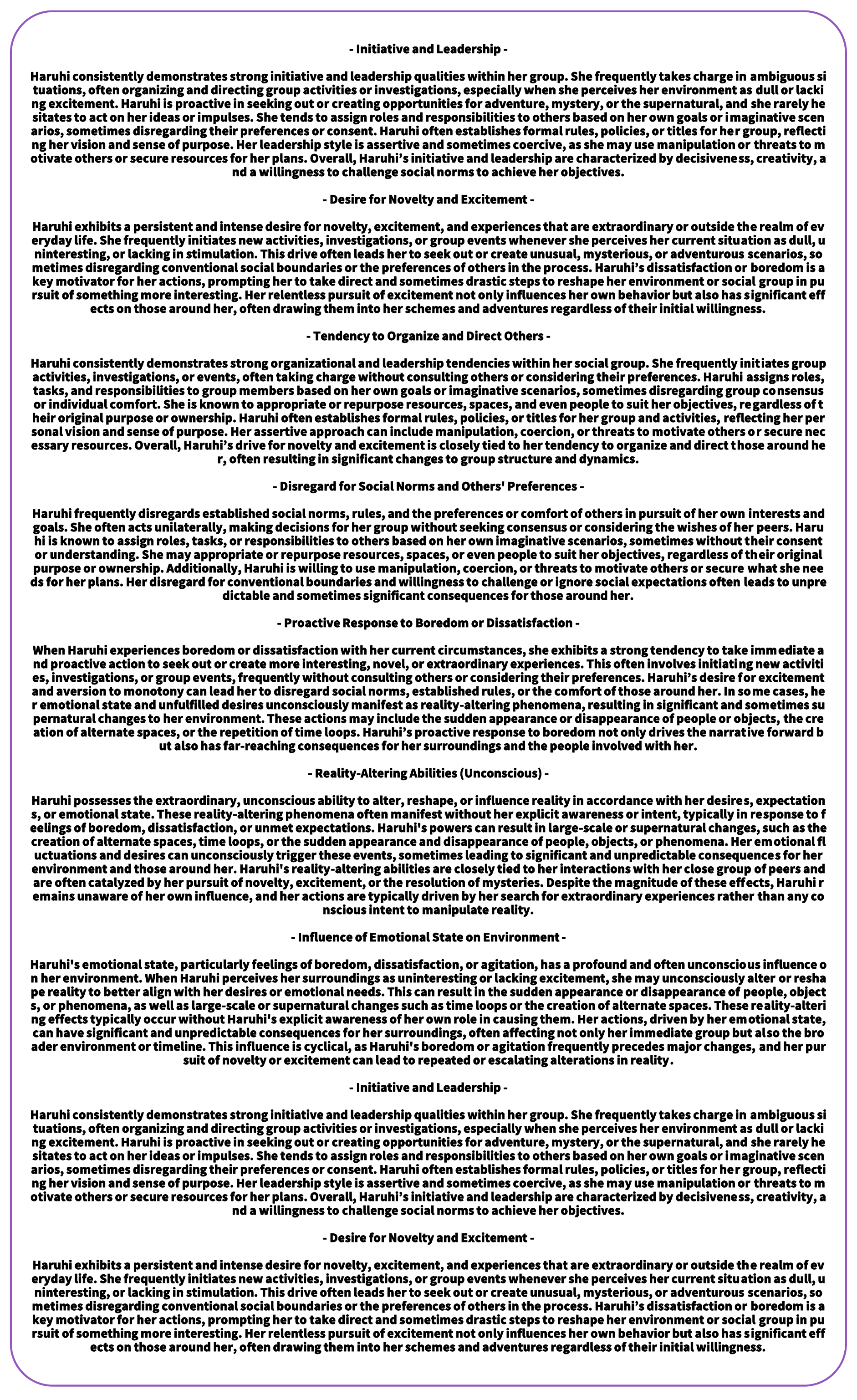}
    \vspace{-3mm}
    \caption{Full wikified CDT for \textit{``Haruhi Suzumiya''} (1/3)}
    \label{fig:full_wikified_cdt_1}
    \vspace{-5mm}
\end{figure*}

\begin{figure*}
    \centering
    \includegraphics[width=0.9\linewidth]{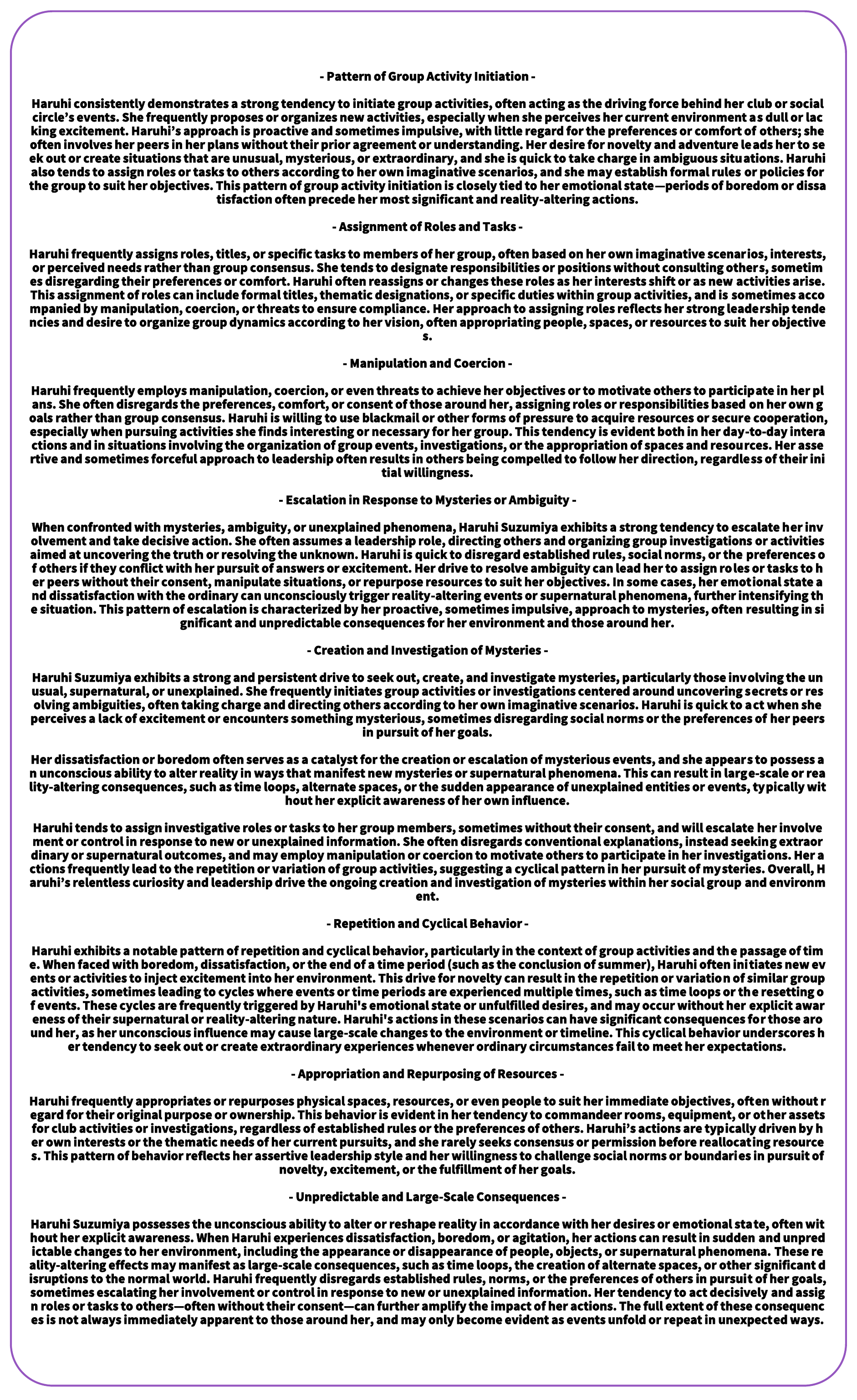}
    \vspace{-3mm}
    \caption{Full wikified CDT for \textit{``Haruhi Suzumiya''} (2/3)}
    \label{fig:full_wikified_cdt_2}
    \vspace{-5mm}
\end{figure*}

\begin{figure*}
    \centering
    \includegraphics[width=0.9\linewidth]{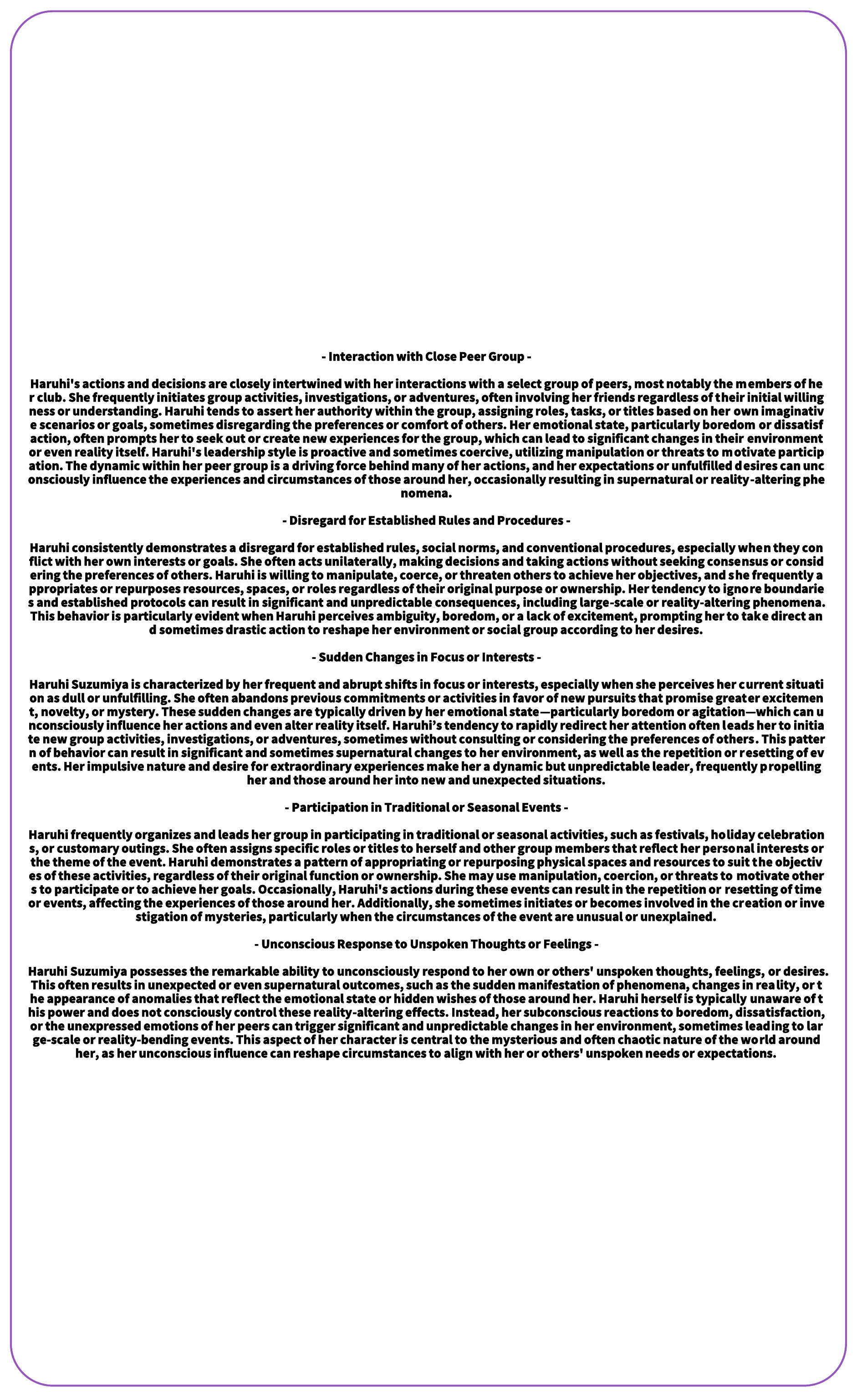}
    \vspace{-3mm}
    \caption{Full wikified CDT for \textit{``Haruhi Suzumiya''} (3/3)}
    \label{fig:full_wikified_cdt_3}
    \vspace{-5mm}
\end{figure*}

\clearpage

\section{Prompts and Templates}

\begin{figure*}
    \centering
    \includegraphics[width=0.9\linewidth]{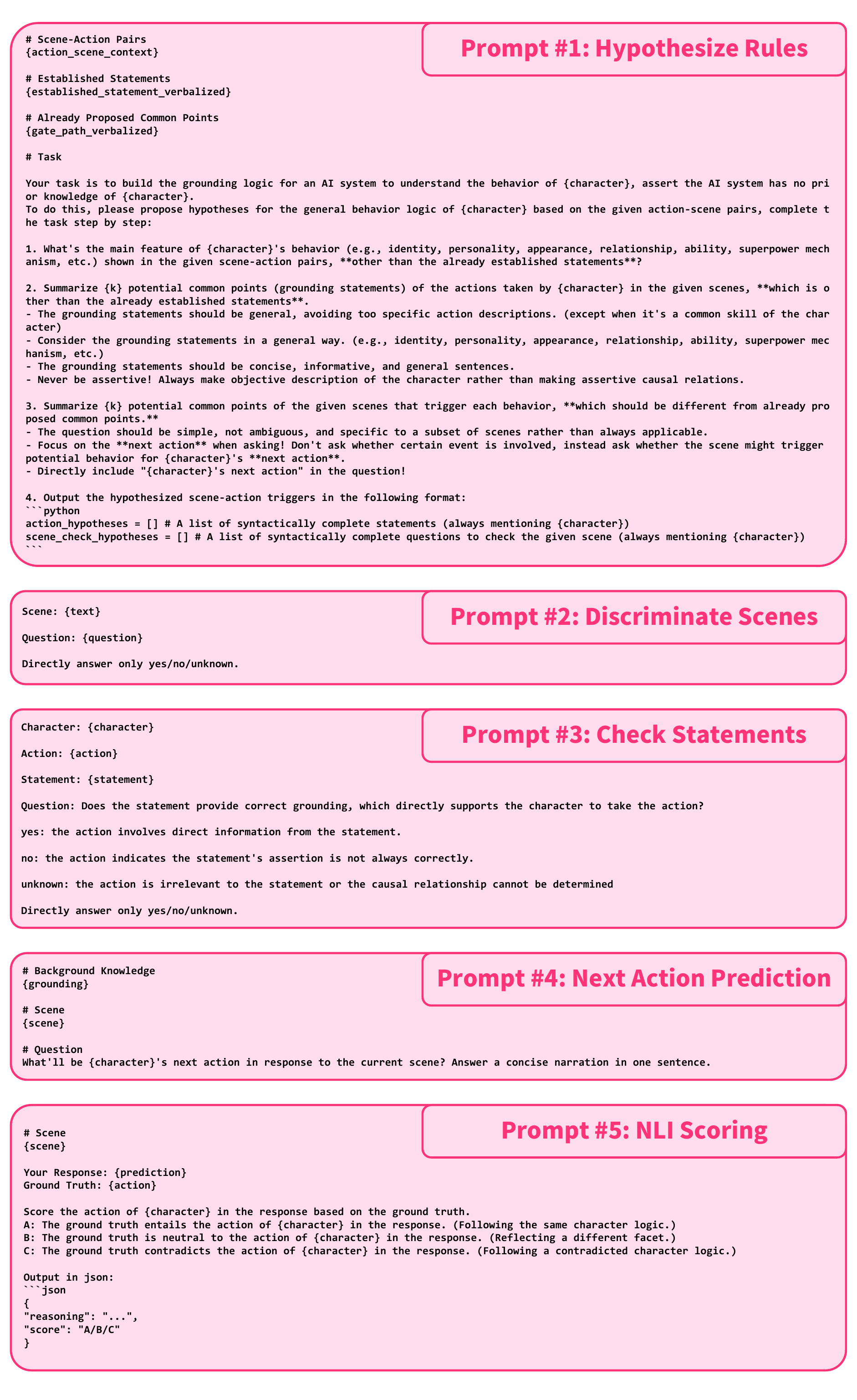}
    \vspace{-3mm}
    \caption{Prompts/Templates used in the main experiments for CDT.}
    \label{fig:main_prompts}
    \vspace{-5mm}
\end{figure*}

\begin{figure*}
    \centering
    \includegraphics[width=0.9\linewidth]{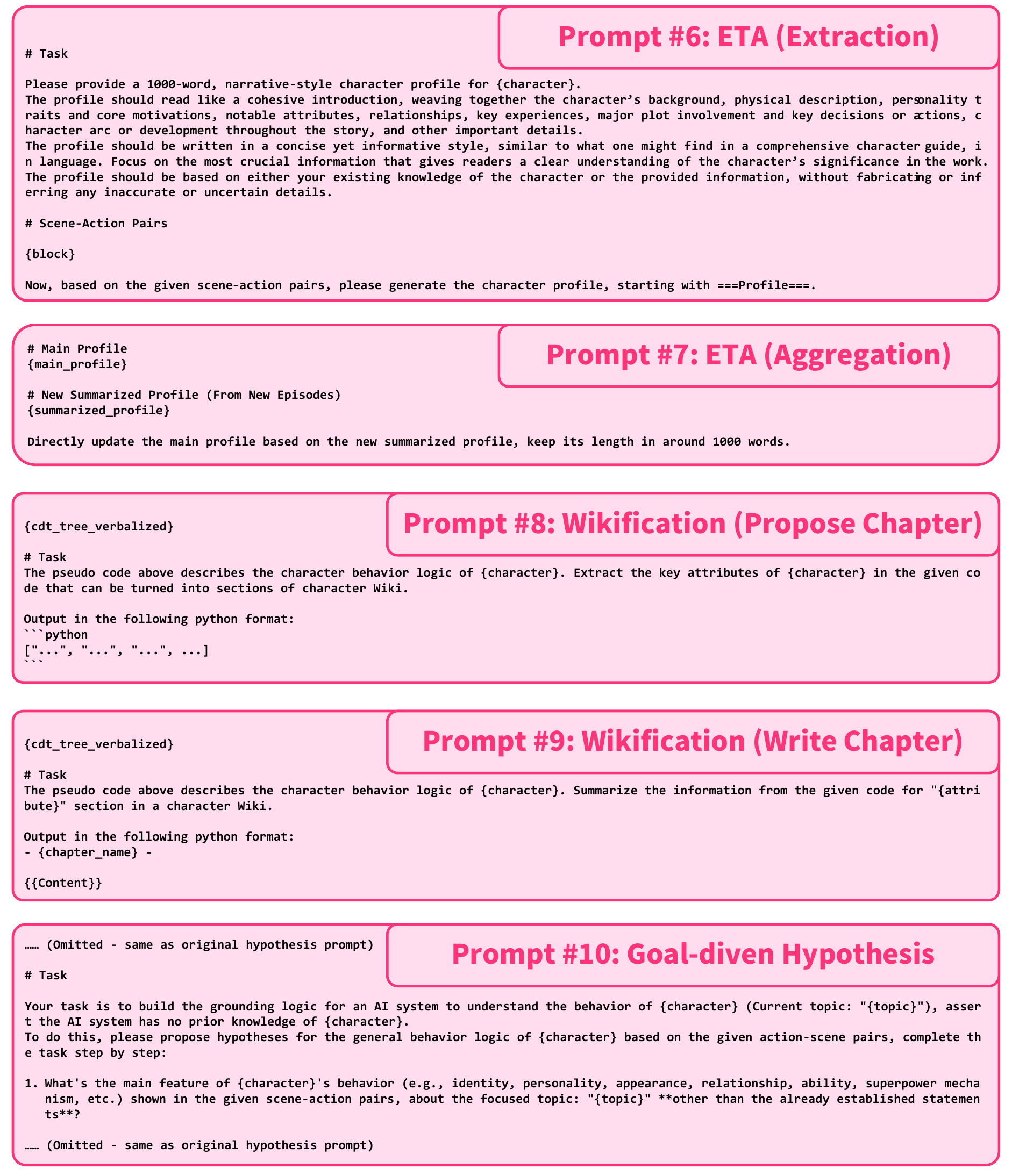}
    \vspace{-3mm}
    \caption{Prompts/Templates used for baselines and variants.}
    \label{fig:baseline_prompts}
    \vspace{-5mm}
\end{figure*}

\begin{figure*}
    \centering
    \includegraphics[width=0.9\linewidth]{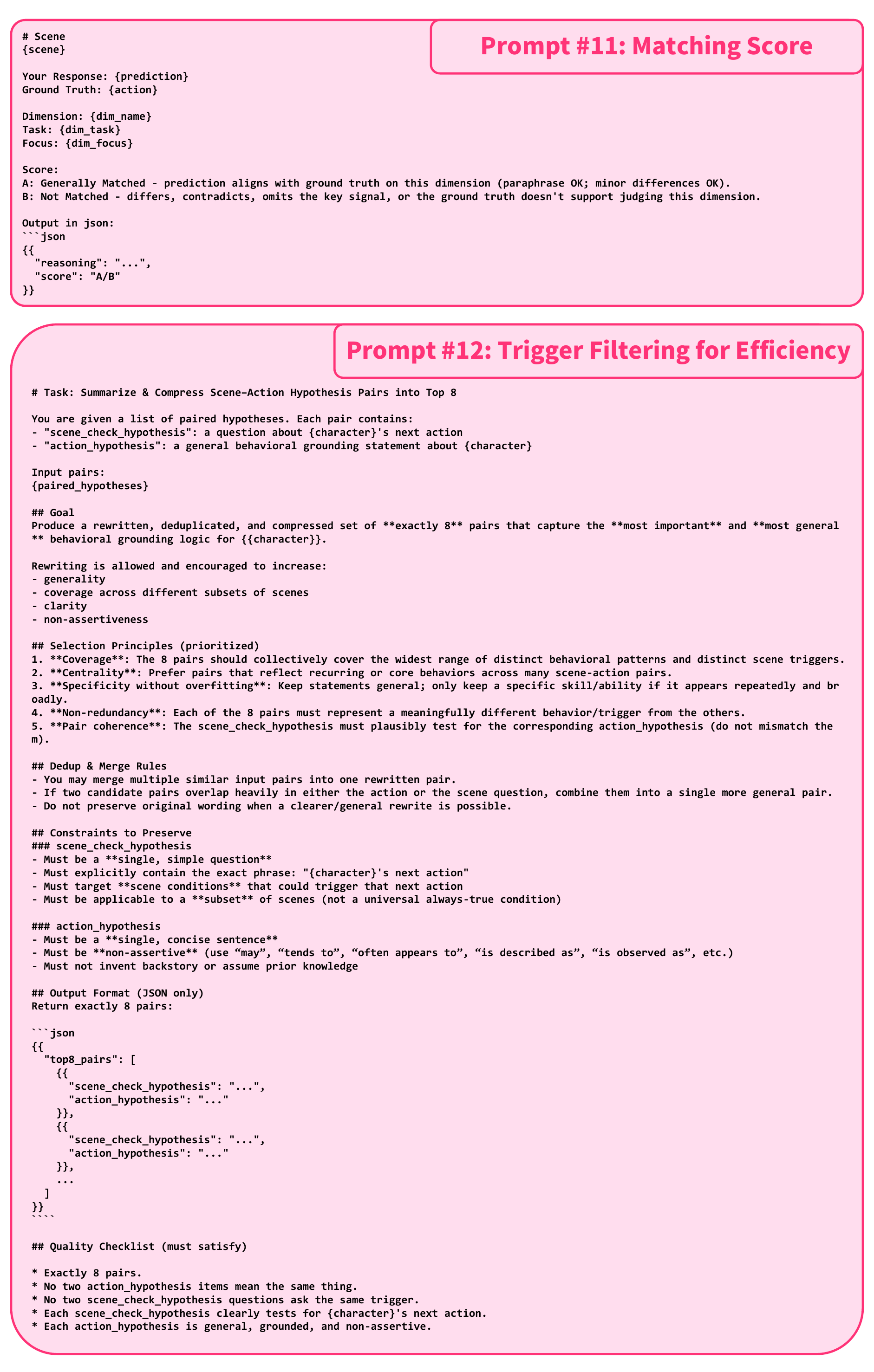}
    \vspace{-3mm}
    \caption{Prompts/Templates used for experiments in appendices.}
    \label{fig:appendix_prompts}
    \vspace{-5mm}
\end{figure*}

We place the prompts used in the main experiment and analyses in Figures~\ref{fig:main_prompts},~\ref{fig:baseline_prompts}, and~\ref{fig:appendix_prompts} for result reproduction.

\clearpage

\section{Character \& Artifact Background Information}
\label{apdx:character_info}

We place concise descriptions of artifacts and characters used in our experiments from Table~\ref{tab:story_info} to~\ref{tab:band_character_info}.

\begin{table*}
\centering
\small
\scalebox{.9}{
\begin{tabular}{lp{11.5cm}}
\toprule
\textbf{Artifact} & \textbf{Concise Abstract} \\
\midrule

The Melancholy of Haruhi Suzumiya & 
A fast-paced school comedy–mystery in which the whims of the eccentric Haruhi unknowingly distort reality. 
Everyday club activities are interwoven with supernatural anomalies, while Kyon’s pragmatic narration provides stability amid escalating chaos. \\
\midrule

K-On! & 
A gentle, music-focused slice-of-life narrative portraying the everyday experiences of the Light Music Club. 
Rather than plot-driven conflict, the story highlights friendship, routine, and subtle personal growth through shared musical practice and leisure. \\
\midrule

Fullmetal Alchemist & 
A serious fantasy adventure following two brothers who turn to alchemy to reclaim what they lost in a forbidden ritual. 
The story blends action with ethical dilemmas, examining themes of sacrifice, human value, and the consequences of power within a turbulent political world. \\
\midrule

JoJo’s Bizarre Adventure (Part 3) & 
A worldwide supernatural journey in which Jotaro and his companions battle enemies using Stand abilities. 
The narrative is defined by inventive combat, strategic confrontations, and flamboyant character interactions that balance intensity with humor. \\
\midrule

Spy $\times$ Family & 
A genre-blending family comedy revolving around a secret agent who forms a fake household to complete a covert mission. 
Espionage, action, and humor coexist with heartfelt moments, as hidden identities collide with genuine emotional bonds. \\
\midrule

Death Note & 
A psychological cat-and-mouse thriller in which a gifted student gains the power to kill by writing names in a supernatural notebook. 
The narrative explores justice, morality, and ego through intense intellectual duels between equally brilliant adversaries. \\

\midrule

A Game of Thrones & 
A large-scale political fantasy centered on competing noble houses locked in cycles of alliance, betrayal, and warfare. 
Personal ambition and moral uncertainty unfold against the backdrop of an approaching existential threat beyond human conflicts. \\
\midrule

Avatar: The Last Airbender & 
An epic coming-of-age tale following Aang as he learns to master elemental powers to restore balance to the world. 
The series combines humor and action with emotional development, emphasizing responsibility, forgiveness, and personal identity. \\

\bottomrule
\end{tabular}
}
\vspace{2mm}
\caption{Concise story descriptions of artifacts used in our experiments (Fine-grained Fandom Benchmark part).}
\label{tab:story_info}
\end{table*}

\begin{table*}
\centering
\small
\scalebox{.9}{
\begin{tabular}{lp{11.5cm}}
\toprule
\textbf{Band} & \textbf{Concise Description} \\
\midrule

Poppin’Party & 
A bright, guitar-driven pop-rock band formed by high school friends, defined by upbeat melodies and an earnest, forward-chasing spirit. 
Their story emphasizes friendship, first dreams, and the steady growth that comes from practicing together and performing as a team. \\
\midrule

Afterglow & 
A straight-ahead rock band built on long-standing childhood bonds, carrying a raw, livehouse energy and a no-frills attitude toward music. 
Their narrative centers on loyalty, everyday honesty, and the tension between staying the same and growing up without breaking the group’s core. \\
\midrule

Pastel*Palettes & 
An idol-style unit whose polished, cheerful image is sustained by behind-the-scenes effort, discipline, and constant on-the-job learning. 
The band’s arc focuses on professionalism, teamwork under pressure, and the contrast between staged perfection and genuine personal struggle. \\
\midrule

Roselia & 
A gothic, high-intensity band pursuing a “perfect” sound through rigorous practice, strong ambition, and uncompromising standards. 
Their storyline highlights artistic pride, conflict born from high expectations, and the hard-won trust required to perform at the highest level. \\
\midrule

Hello, Happy World! & 
A flamboyant, joy-first band that treats performance as a mission to make the world smile, blending showmanship with playful chaos. 
Comedic set pieces coexist with sincere warmth, as the members’ eccentricities ultimately reinforce a shared commitment to spreading happiness. \\
\midrule

Morfonica & 
A melodic rock group distinguished by the prominence of violin, combining refined textures with youthful sensitivity and introspective emotion. 
Their narrative explores confidence, talent gaps, and self-acceptance, as the band learns to transform insecurity into a cohesive musical identity. \\
\midrule

RAISE A SUILEN & 
A hard-hitting rock/electronic hybrid centered on precision, speed, and stage dominance, built through deliberate recruitment and relentless rehearsal. 
The story foregrounds professionalism, creative control, and the friction—and eventual cohesion—of strong personalities striving for the same peak. \\
\midrule

MyGO!!!!! & 
A volatile, emotion-forward rock band whose sound is shaped by conflict, vulnerability, and the members’ difficulty with honesty and connection. 
Their arc focuses on miscommunication, fragile belonging, and the intense catharsis of turning personal pain into music and mutual commitment. \\

\bottomrule
\end{tabular}
}
\vspace{2mm}
\caption{Concise band descriptions used in our experiments (Bandori Conversational Benchmark part).}
\label{tab:band_info}
\end{table*}

\begin{table*}
\centering
\small
\scalebox{.75}{
\begin{tabular}{cccp{18cm}}
\toprule
\multirow{5}*{\rotatebox{90}{Haruhi}} & \multirow{5}*{} & Haruhi & An impulsive, hyperactive high school girl whose restless curiosity and odd worldview trigger the story’s cascade of bizarre events. \\
&  & Kyon & A sardonic, level-headed student who narrates events and acts as Haruhi Suzumiya’s reluctant yet stabilizing partner. \\
&  & Nagato & A silent, unreadable SOS Brigade member marked by extraordinary intellect and mysterious, otherworldly roots. \\
&  & Koizumi & An always-smiling transfer student and esper who aids the Brigade while carefully guarding critical secrets. \\
&  & Asahina & A timid, kind upperclassman conscripted into the SOS Brigade as their cute, enigmatic “mascot,” frequently dragged into their antics. \\
\midrule
\multirow{5}*{\rotatebox{90}{K-On!}} & \multirow{5}*{} & Yui & The bubbly, scatterbrained lead guitarist of the light music club, whose boundless energy—and sweet tooth—keeps the band moving. \\
&  & Ritsu & The boisterous, prank-loving drummer whose playful antics and casual leadership keep the group upbeat and united. \\
&  & Mio & A shy but highly capable bassist, gentle at heart and blessed with sharp musical sensitivity. \\
&  & Mugi & Tsumugi Kotobuki, a kind, affluent keyboardist who loves pampering her friends and making club life feel luxurious. \\
&  & Azusa & A hardworking, gifted junior guitarist who soon becomes essential to the club’s tight sound and practice habits. \\
\midrule
\multirow{5}*{\rotatebox{90}{FMA}} & \multirow{5}*{} & Edward & A gifted, stubborn young alchemist who journeys to recover his and his brother’s bodies after a catastrophic transmutation. \\
&  & Alphonse & A gentle, big-hearted boy whose soul dwells in a hulking suit of armor, traveling with his brother to regain what they lost. \\
&  & Winry & A talented automail mechanic and the Elrics’ childhood friend, renowned for her technical skill and steadfast compassion. \\
&  & Roy & A charismatic, driven State Alchemist and master of flame, intent on reshaping the military from the inside. \\
&  & Ling & A charismatic, relentless prince from Xing who pursues immortality while carrying a heavy duty to his nation. \\
\midrule
\multirow{7}*{\rotatebox{90}{JOJO}} & \multirow{7}*{} & Jotaro & A stoic, seemingly unshakable high schooler and “Stardust Crusaders” lead, famed for Star Platinum and iron resolve. \\
&  & Polnareff & A bold, flamboyant French swordsman who allies with the Crusaders, fighting through the swift Stand Silver Chariot. \\
&  & Joseph & A fast-thinking, over-the-top Joestar whose schemes and bravado—“Your next line is…”—repeatedly flip battles in his favor. \\
&  & DIO & A magnetic, utterly ruthless vampire whose towering ambition and cry of “Za Warudo!” cement him as a legendary foe. \\
&  & Kakyoin & A composed, analytic ally in “Stardust Crusaders,” battling with Hierophant Green, a Stand that attacks with emerald blasts. \\
&  & Avdol & A wise, steadfast Egyptian Stand user whose Magician’s Red commands fierce flames and unshakable backing. \\
&  & Iggy & A grumpy Boston Terrier Stand user with a fondness for coffee gum, whose reluctant heroics turn out to be vital. \\
\midrule
\multirow{5}*{\rotatebox{90}{DN}} & \multirow{5}*{} & Light & A brilliant, idealistic student who acquires the Death Note and resolves to reshape the world through absolute, lethal justice. \\
&  & L & An eccentric, reclusive genius detective whose unconventional methods and sharp intuition pit him directly against Kira. \\
&  & Near & A calm, analytical prodigy who succeeds L, relying on detached logic and meticulous planning to pursue the truth. \\
&  & Misa & A devoted idol and second Kira, driven by love and gratitude, whose impulsive loyalty complicates the deadly mind games. \\
&  & Mello & A volatile, fiercely competitive successor to L who embraces risk and criminal alliances to outmaneuver his rivals. \\
\midrule
\multirow{3}*{\rotatebox{90}{S$\times$F}} & \multirow{3}*{} & Loid & An elite undercover agent who assembles a fake family for a high-stakes mission, balancing espionage with improvised parenthood. \\
&  & Yor & A soft-spoken civil servant secretly working as a lethal assassin, struggling to reconcile her double life with domestic normalcy. \\
&  & Anya & A cheerful, telepathic child who knows everyone’s secrets, holding the family together through innocence and quiet insight. \\
\midrule
\multirow{11}*{\rotatebox{90}{AGOT}} & \multirow{11}*{} & Tyrion & The razor-witted youngest Lannister, Tyrion navigates Westerosi politics with wit, nerve, and dark humor despite a lifetime of scorn for his size. \\
&  & Daenerys & An exiled Targaryen princess who starts as a hesitant pawn and evolves into a determined, power-claiming ruler. \\
&  & Cersei & An ambitious, scheming queen whose beauty conceals a ruthless devotion to her family and grip on power. \\
&  & Jaime & The notorious Kingslayer—charming, deadly, and deeply conflicted—whose sworn duties and loyalties are tangled and fraught. \\
&  & Robb & The dutiful heir of Winterfell, pushed too soon into command and responsibility by his family’s misfortune. \\
&  & Eddard & The resolute Lord of Winterfell, a man of stern honor who serves as Warden of the North. \\
&  & Arya & A fiercely independent Stark girl who casts off courtly roles in favor of freedom, training, and the blade. \\
&  & Catelyn & The determined Lady of Winterfell, driven by fierce maternal loyalty and a firmly practical mind. \\
&  & Sansa & The elder Stark daughter, cherished for grace and manners, whose romantic dreams collide with brutal reality. \\
&  & Jon & Eddard’s brooding illegitimate son, raised at Winterfell and driven by questions of identity, duty, and quiet resolve. \\
&  & Bran & A curious young Stark whose devastating fall thrusts him onto an unforeseen and fateful journey. \\
\midrule
\multirow{4}*{\rotatebox{90}{ATLA}} & \multirow{4}*{} & Aang & The final Airbender and hesitant Avatar, playful at heart yet burdened with restoring balance to a world in war. \\
&  & Katara & A determined, compassionate waterbender from the Southern Tribe who grounds the group and refuses to tolerate injustice. \\
&  & Sokka & A wisecracking, inventive warrior whose boomerang skills and ingenuity repeatedly end up saving the day. \\
&  & Zuko & An exiled Fire Nation prince, driven by a burning quest for honor that gradually turns into a search for a new self. \\
\bottomrule
\end{tabular}
}
\vspace{2mm}
\caption{Simple background information of characters in our experiments (Fandom Benchmark part).}
\label{tab:character_info_fandom}
\end{table*}

\begin{table*}
\centering
\small
\scalebox{.75}{
\begin{tabular}{cccp{18cm}}
\toprule
\multirow{5}*{\rotatebox{90}{PoPiPa}} & \multirow{5}*{} & Kasumi & An upbeat, starry-eyed vocalist–guitarist whose impulsive enthusiasm pulls people together and kicks off the band’s journey. \\
&  & Tae & A free-spirited lead guitarist with strong technique and quirky instincts, often drifting at her own pace yet boosting the band’s sound. \\
&  & Rimi & A shy, gentle bassist who grows braver through performance, bringing careful support and warm sincerity to the group. \\
&  & Saaya & A dependable drummer with a caring, family-first mindset, acting as the band’s steady backbone in both practice and life. \\
&  & Arisa & A sharp-tongued but reliable keyboardist whose practicality and quick thinking keep the band organized, grounded, and moving forward. \\
\midrule

\multirow{5}*{\rotatebox{90}{AG}} & \multirow{5}*{} & Ran & A blunt, prideful vocalist–guitarist who values authenticity, carrying the band’s straightforward rock spirit and stubborn resolve. \\
&  & Moca & A laid-back lead guitarist with a mischievous streak, masking keen observation and musical confidence behind casual teasing. \\
&  & Himari & A bright, encouraging bassist and nominal leader, energizing the group with optimism while trying to hold everyone together. \\
&  & Tomoe & A reliable, big-sister drummer who supports others through calm strength, stepping up whenever the band needs stability. \\
&  & Tsugumi & A kind keyboardist with a gentle, practical touch, often mediating tensions and keeping the group’s everyday rhythm intact. \\
\midrule

\multirow{5}*{\rotatebox{90}{PasuPare}} & \multirow{5}*{} & Aya & A relentlessly earnest vocalist who chases the idol dream through effort and persistence, learning confidence by doing the work. \\
&  & Hina & A cheerful, genius guitarist who loves “fun” above all, acting on bright ideas with little hesitation and lots of momentum. \\
&  & Chisato & A cool, realistic bassist with strong professionalism, frequently reining in chaos while protecting the group’s long-term direction. \\
&  & Maya & A drummer with deep audio-gear passion and technical know-how, becoming animated when music setups and stage craft are involved. \\
&  & Eve & A sincere keytarist devoted to “bushido,” whose wholehearted intensity and kindness can be both inspiring and unexpectedly disruptive. \\
\midrule

\multirow{5}*{\rotatebox{90}{Roselia}} & \multirow{5}*{} & Yukina & A fiercely driven vocalist who pursues a “perfect” sound, pushing herself and others with uncompromising standards and focus. \\
&  & Sayo & A serious, disciplined guitarist who relies on hard work over flair, expressing care through responsibility and relentless practice. \\
&  & Lisa & A warm, attentive bassist who acts as the band’s emotional glue, balancing high ambition with everyday empathy and reassurance. \\
&  & Ako & A high-energy drummer with a dramatic, chuuni-tinged flair, bringing loud confidence while still craving recognition and growth. \\
&  & Rinko & A shy, soft-spoken keyboardist with exceptional skill, gradually building courage through supportive bonds and shared performances. \\
\midrule

\multirow{5}*{\rotatebox{90}{HHW}} & \multirow{5}*{} & Kokoro & A wealthy, fearless optimist who treats making people smile as a mission, turning wild ideas into surprisingly sincere action. \\
&  & Kaoru & A theatrical guitarist who plays the “prince” role with flourish, using charm and melodrama to lift the mood around her. \\
&  & Hagumi & A sunny, energetic bassist with an athletic, straightforward vibe, often charging ahead with honest excitement and big smiles. \\
&  & Kanon & A timid but kind drummer who constantly pushes past fear, finding bravery through small steps and friends who believe in her. \\
&  & Misaki & A pragmatic, overworked coordinator (and DJ) who keeps the group functional, often acting as the lone realist amid cheerful chaos. \\
\midrule

\multirow{5}*{\rotatebox{90}{Monica}} & \multirow{5}*{} & Mashiro & A sensitive vocalist and lyricist who struggles with insecurity, slowly learning to voice her feelings through song and companionship. \\
&  & Touko & A flashy, extroverted lead guitarist who loves attention and momentum, bringing brightness while occasionally stirring trouble by impulse. \\
&  & Nanami & A multi-talented bassist fixated on being “normal,” masking inner conflict with humor and adaptability across many situations. \\
&  & Tsukushi & A hardworking drummer and leader who tries to be dependable, persisting through clumsiness with determination and care for the team. \\
&  & Rui & A cool, perfection-driven violinist and composer who prioritizes results, gradually confronting the role of emotion and trust in music. \\
\midrule

\multirow{5}*{\rotatebox{90}{RAS}} & \multirow{5}*{} & CHU$^2$ & A demanding genius DJ/producer who builds the band with strict control and ambition, driving everyone toward a professional-level stage. \\
&  & LAYER & A sharp, charismatic bassist–vocalist whose powerful presence and steady musicianship anchor the band’s sound under intense expectations. \\
&  & LOCK & A young, earnest guitarist who grows through pressure and mentorship, balancing admiration with the need to prove her own worth. \\
&  & MASKING & A fearless, high-impact drummer who thrives on adrenaline and volume, powering performances with wild confidence and physical intensity. \\
&  & PAREO & A devoted keyboardist with a shy core and idol-like polish, channeling loyalty and effort into supporting the band’s vision. \\
\midrule

\multirow{5}*{\rotatebox{90}{MyGO}} & \multirow{5}*{} & Tomori & A withdrawn, highly sensitive vocalist and lyricist who clings to “words” for connection, turning pain and longing into songs. \\
&  & Anon & A social, image-savvy rhythm guitarist who wants to belong and be seen, learning sincerity as her confident front gets tested. \\
&  & Raana & A freewheeling lead guitarist with a mysterious, playful calm, following curiosity and sound first while ignoring most social rules. \\
&  & Soyo & A gentle, composed bassist who tries to keep harmony, often caught between caring intentions and the pressure of unresolved history. \\
&  & Taki & A blunt, intense drummer and composer whose strict standards hide protectiveness, expressing concern through sharp honesty and persistence. \\
\bottomrule
\end{tabular}
}
\vspace{2mm}
\caption{Simple background information of characters in our experiments (Bandori Benchmark part).}
\label{tab:band_character_info}
\end{table*}

\section{LLM Usage Statement}
We used large language models (LLMs) only for copy-editing: improving grammar, clarity, and overall writing style to enhance readability. The LLMs did not contribute to or modify the scientific content, core ideas, methodology, or experimental results. The authors take full responsibility for the manuscript’s final content and accuracy.

\end{document}